\documentclass{article}

\usepackage[preprint]{neurips_2024}

\usepackage[utf8]{inputenc} 
\usepackage[T1]{fontenc}    
\usepackage{hyperref}       
\usepackage{url}            
\usepackage{booktabs}       
\usepackage{amsfonts}       
\usepackage{nicefrac}       
\usepackage{microtype}      
\usepackage{xcolor}         
\usepackage{graphicx}
\usepackage{subcaption}
\usepackage{wrapfig}
\usepackage{natbib}
\usepackage{float}
\usepackage{listings}

\setcitestyle{numbers}
\setcitestyle{square}
\begin{document}

\title{Uncovering Layer-Dependent Activation Sparsity Patterns in ReLU Transformers}
%

\author{
  Cody Wild\thanks{Corresponding author} \\
  Google Research \\
  \texttt{codywild@google.com}
  \And
  Jesper Anderson \\
  Google Research \\
  \texttt{jespera@google.com}
}

\maketitle
\begin{abstract}
  Previous work has demonstrated that MLPs within ReLU Transformers exhibit high levels of sparsity, with many of their activations equal to zero for any given token. We build on that work to more deeply explore how token-level sparsity evolves over the course of training, and how it connects to broader sparsity patterns over the course of a sequence or batch, demonstrating that the different layers within small transformers exhibit distinctly layer-specific patterns on both of these fronts. In particular, we demonstrate that the first and last layer of the network have distinctive and in many ways inverted relationships to sparsity, and explore implications  for the structure of feature representations being learned at different depths of the model. We additionally explore the phenomenon of ReLU dimensions "turning off", and show evidence suggesting that "neuron death" is being primarily driven by the dynamics of training, rather than simply occurring randomly or accidentally as a result of outliers.
\end{abstract}

\section{Introduction}
\begin{wrapfigure}{r}{.6\linewidth}
  \centering
  \includegraphics[width=\linewidth]{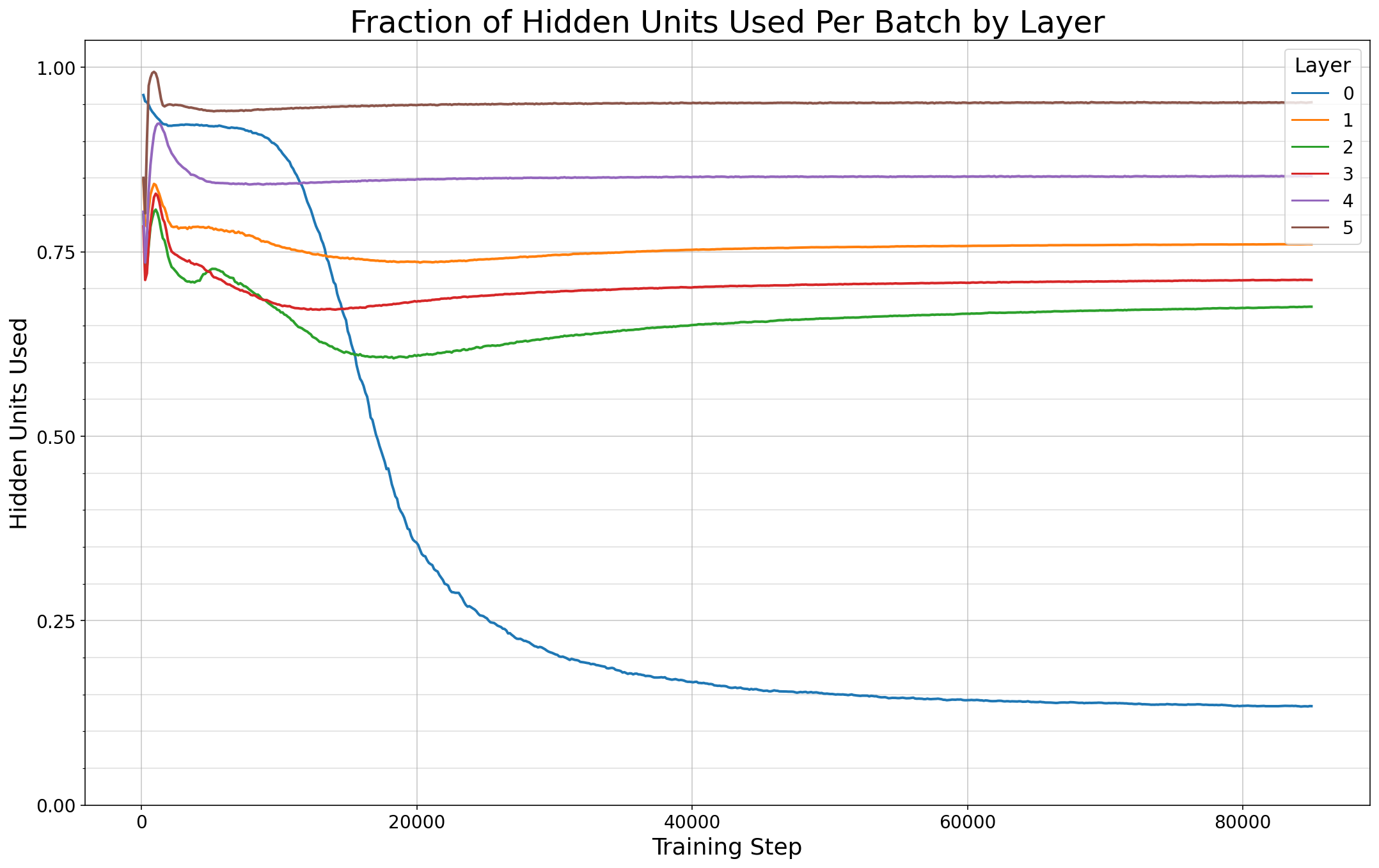}
  \caption{The fraction of hidden units in the MLPs of a six-layer Transformer that have a nonzero activation in the course of a 8192-token batch, divided by layer, shown over the course of training. We can see there here that there is strong divergence in the behavior of different layers, and that in particular the first layer collapses dramatically in how many of its available hidden units it uses}
  \label{batch_dims}
\end{wrapfigure}

In deep networks, hard-zero sparsity  can be a useful property---both for understanding the computations being made with more clarity, and for potentially making those computations faster than they could be otherwise \citep{mirzadeh2023relu}. This work focuses specifically on the activation dynamics of the two-layer MLPs in a standard Transformer block \citep{vaswani2023attention}. While \citet{li2023lazy} demonstrated that trained networks using ReLU nonlinearities exhibit sparsity on a per-token basis, we believe this is the first work to examine in depth how sparsity evolves over training, both on a per-token level and between tokens in a sequence or batch.
\\
\\

\textbf{We show evidence that:}
\begin{itemize}
    \item \textbf{The extent of sparsity, and also its evolution over the course of training, follow dramatically different patterns based on layer index}, with the first layer and final layer in the network occupying distinct and opposing niches in terms of their sparsity behavior. 
    \item \textbf{Sparsity patterns at the token, sequence, and batch of sequences frequently anticorrelate}, with the layers that have the activate the fewest hidden units per-token frequently being those that use the highest number over the course of a sequence or batch. 
    \item \textbf{Many hidden units are ``turned off'' during training -- never used in a batch, and of empirically low value to subsequent training -- particularly in the first layer of the network.} We demonstrate that the dynamics of this ``neuron death'' vary dramatically by layer, with some layers gaining neurons over training while others lose them. Furthermore, \textbf{neuron death only occurs in specified regimes of training, suggesting it is related to learning dynamics dynamics}, rather than simply an accidental process driven by a neuron being "pushed over the threshold" in a single batch. 
    \item While the majority of these never-used hidden units turn off during the course of training, \textbf{a small fraction - about 5\% of the total hidden dimension of a layer - are at or below zero for every token at initialization} and never turn on. This is suggestive that it might be possible to better optimize initializations used for ReLU transformers, achieving higher effective use of MLP hidden dimensions, although this work does not directly test this. 

\end{itemize}

\section{Methodology}
\subsection{Metrics}
\label{metrics_discussion}
While attention mechanisms are the feature that differentiate Transformers from other sequence models, the Multi-Layer Perceptrons (MLPs) within a Transformer block seem to be key to how they learn and store information. MLPs contain the majority of parameters within most modern language models, and it is believed that they are central to how models store factual information\cite{meng2023locating}. Given this, we believe that rigorous work to better understand the internal machinery of Transformer MLPs is of potentially high value to our understanding of this important class of models. 

Previous work demonstrated that MLPs that use ReLUs tend to be sparse in their activations - that is, for any given token vector that makes a forward pass through a MLP, only a small fraction of the available hidden units will have values greater than 0 after a ReLU activation is applied \citep{li2023lazy}. This work additionally focuses on sequence and batch measures of sparsity. We believe these metrics are important both for understanding the degree to which sparsity can or can't be computationally exploited, and also for understanding what sparsity might be able to tell us about features being learned. 

To investigate this, we define and calculate three core metrics of sparsity - or, inversely, of ``hidden unit use''. (We decided that focusing on the positive measure was more intuitive in this setting, but this is fundamentally just sparsity from a different perspective). These metrics are calculated on the hidden dimension activations between the two dense layers that make up a Transformer MLP, and all of them are calculated independently for each layer in the network.
\begin{enumerate}
    \item \textbf{Per-Token Hidden Unit Use} - The number of hidden units that are nonzero post-ReLU, calculated per token, and then averaged over sequence and batch. 
    \item \textbf{Per-Sequence Hidden Unit Use} - The number of hidden units that are nonzero post-ReLU \textit{anywhere in a sequence}. This is fundamentally an ``or'' over all token in a sequence - if a hidden unit is used in even one token, it is included in this count. Experiments in this paper are done with a standard sequence length of 512. 
    \item \textbf{Per-Batch Hidden Unit Use} - The number of hidden units that are nonzero post-ReLU anywhere in a batch. This is similar to Sequence Hidden Unit Use in that a hidden unit dimension is counted here if any token in any sequence in the batch has a value greater than zero. Metrics are calculated per-device, with a device batch size of 16, meaning a batch for this purpose contains 16*512 = 8192 tokens.
\end{enumerate}

We additionally calculate percentile metrics of how frequently over the course of a sequence a hidden unit neuron is used, when it is used at all. These are calculated by first taking the hidden units that have a nonzero value anywhere in the current sequence, and, for each of them, counting what fraction of tokens in that sequence each hidden unit is used on. We then take different percentiles of that distribution. This can intuitively be thought of as taking a histogram of how frequently in a sequence a hidden unit is used, removing values of zero, and taking different percentiles of that histogram, for the sake of making a metric that can aggregate more easily over training. 

For all the plots shown in this paper, we chose to focus on the early stages of training, since behavior tends to stabilize by about 70,000 steps, and showing a full plot makes the early-in-training dynamics harder to see. Full plots are available in the appendix whenever a shortened version is shown.

\subsection{Model Architecture} 
The primary subject of investigation in this paper is a 6-layer decoder-only Transformer model, with a hidden dimension size of 32768 and ReLU MLP nonlinearity, trained on C4 next token prediction \citep{c4citation}. LeCun Normal initialization is used for the dense kernels within the MLPs. An AdamW optimizer is used, with a cosine decay learning rate schedule after a warm-up of 5000 steps up to a peak LR of 3e-3. In section \ref{hyperparameter_tests} we discuss the extent to which our observations transfer to different model depths, hidden dimensions, and learning rates, but outside of that section all results are shown on the model described above. Each model was trained with GCP TPU-v2 slice of type v2-32, and in this environment took about 2.5 days of wallclock time to reach convergence. 

\section{Results}
\begin{table}[]
\centering
\begin{tabular}{ccccccc}
\textbf{Layer} & \textbf{Per-Token (\%)} & \textbf{Per-Sequence (\%)} & \textbf{Per-Batch (\%)} & \textbf{Token/Seq Dims} & \textbf{Seq/Batch Dims} \\
0     & 4.1       & 12.2          & 13.3       & 0.346          & 0.917           \\
1     & 6.8       & \textbf{71.7}          & 79.3      & 0.096          & 0.904           \\
2     & \textbf{7.1}       & 58.8          & 71.0      & 0.121          & 0.828           \\
3     & 6.4       & 54.2          & 73.0      & 0.118          & 0.742           \\
4     & 2.6       & 50.7          & 86.1      & 0.051          & 0.589           \\
5     & 3.0       & 58.0          & \textbf{95.6}      & 0.052          & 0.606                
\end{tabular}
\caption{This table shows the percent of nonzero hidden units per-token, per-sequence, and per-batch at convergence, as well as some ratios of those dimension counts. The first layer uses the fewest dimensions per batch, while the final layer uses the most, but that the rankings between the layers differ between aggregation levels.}
\label{used_at_convergence}
\end{table}

\begin{table}[]
\centering
\begin{tabular}{ccccc}
\textbf{Layer} & \textbf{On Batch 0 (\%)} & \textbf{Turned On (\%)} & \textbf{Turned Off (\%)} & \textbf{Final Use (\%)} \\
0     & 97.0           & .05             & 83.8           & 13.2       \\
1     & 93.2           & 1.7            & 15.6            & 79.3      \\
2     & 90.9           & 2.6            & 22.5            & 71.0      \\
3     & 89.0           & 4.3           & 20.2            & 73.1      \\
4     & 88.2           & 8.5           & 10.6            & 86.1      \\
5     & 87.5           & 12.2           & 3.8            & 95.6     
\end{tabular}
\caption{This table shows the percentage of hidden units that were active (or ``on'', as a shorthand) in the first batch of training, the amount that either became active or inactive over the course of training, and the final percentage active at convergence.}
\label{dims_on_off}
\end{table}

\begin{table}[]
\centering
\begin{tabular}{ccccc}
\textbf{Layer} & \textbf{50th Perc(\%)} & \textbf{65th Perc(\%)} & \textbf{75th Perc(\%)} & \textbf{90th Perc(\%)} \\
0     & 13.5    & 31.3    & 59.2    & 94.6    \\
1     & 1.7     & 2.9     & 4.6     & 40.0    \\
2     & 1.6     & 3.2     & 5.9     & 51.1    \\
3     & 1.2     & 2.3     & 4.0     & 35.2    \\
4     & .5      & .8      & 1.2     & 3.7     \\
5     & .6      & 1.1     & 1.8     & 6.9    
\end{tabular}
\caption{This table is created by calculating, for each hidden unit in a layer, the number of times it appears over the course of a sequence, subsetting to hidden units used at least once in the sequence, and then calculating percentiles of appearance count within that subset. For generality, this count is represented as the percentage of tokens in the sequence in which the Nth percentile dimension is used.}
\label{percentiles_table}
\end{table}

Our most salient top-level results, as shown in Table \ref{used_at_convergence} are that: 
\begin{enumerate}
\item Different layers in a network use very different numbers of hidden units over the course of a batch, and this is not a trivial extension of the per-token value. In fact, there is a moderate (-0.15) negative correlation between token and batch hidden unit use.
\item Specifically, the first layer in the network (layer 0) consistently uses the fewest of its available hidden units in a batch (only about 13.3\%) and the final layer in the network (in this case layer 5) uses the most (95.6\%). 
\end{enumerate}

\subsection{Sparsity at Initialization}
To understand how sparsity develops during training, we need to first understand the sparsity characteristics of an untrained model. In Table \ref{dims_on_off}, we can see that, at initialization, higher layers of the network use monotonically fewer of their dimensions at in the first training batch. Given that weight initialization is shared across layers, this suggests that the transformations being applied even by a random network shift the inputs to each successive MLP in a way that induces sparsity. 

Importantly, we can see that a hidden unit being zero in the first batch of training does not mean that it never has the possibility to shift outside of that regime - in the ``Turned On`` column of \ref{dims_on_off} we see that all layers turn on some hidden units that began training unused. The tendency for layers to "turn on"  hidden units is monotonic in the opposite direction, with higher layers turning on more  hidden units. This could be indicative of the model learning features at higher layers that benefit from a larger number of hidden units, but could also mean that, while all layers would benefit from turning on  hidden units, later layers in the network are more able to do so because of the distribution shift happening underneath them in the earlier layers.\\

\subsection{How Sparsity Evolves Over the Course of Training}
\begin{figure}[h]
\begin{subfigure}{.5\textwidth}
  \centering
\includegraphics[width=\linewidth]{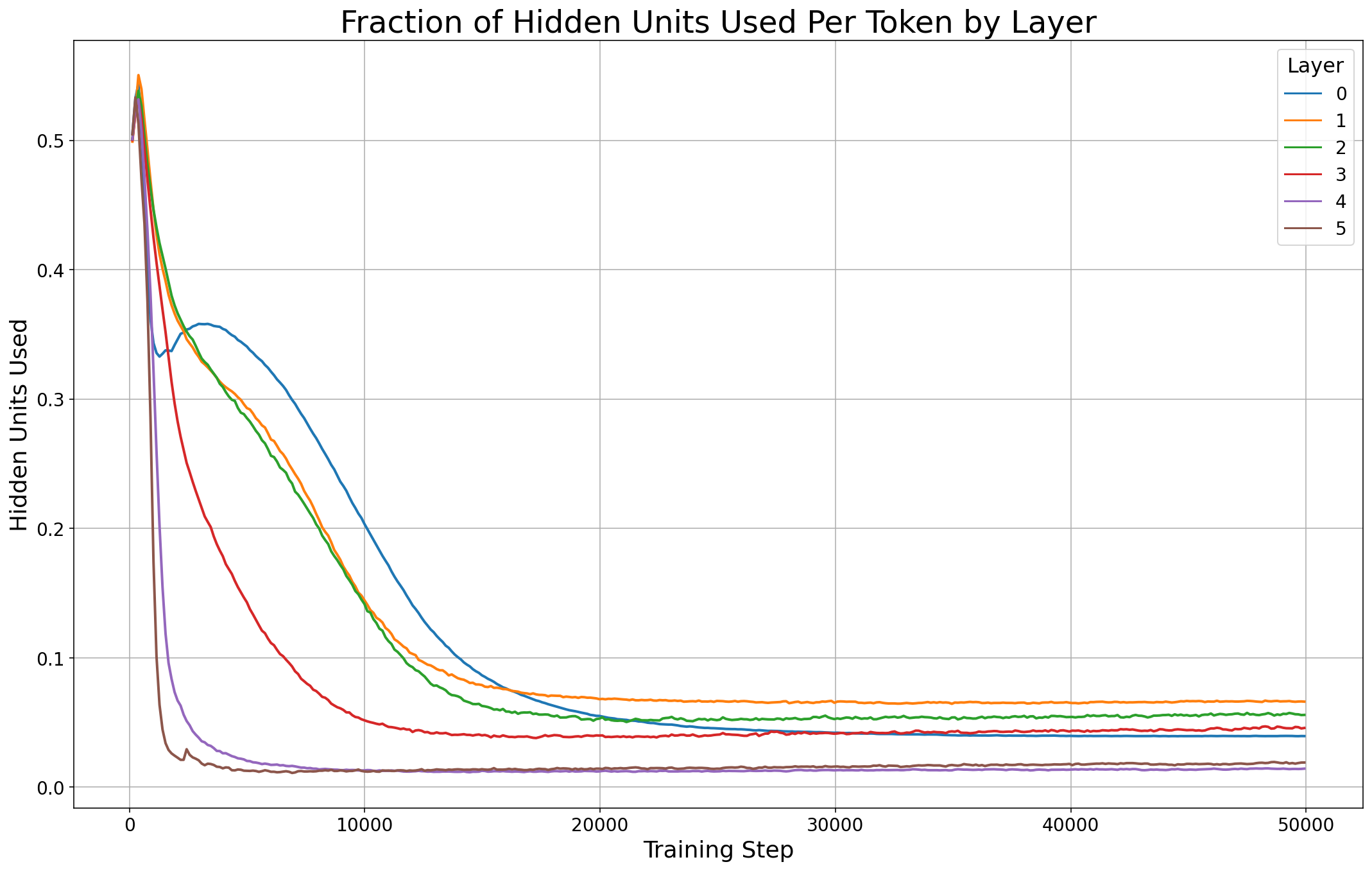}
  \caption{Hidden Unit Use Per Token}
  \label{token_over_training:token}
\end{subfigure}%
\begin{subfigure}{.5\textwidth}
  \centering
  \includegraphics[width=\linewidth]{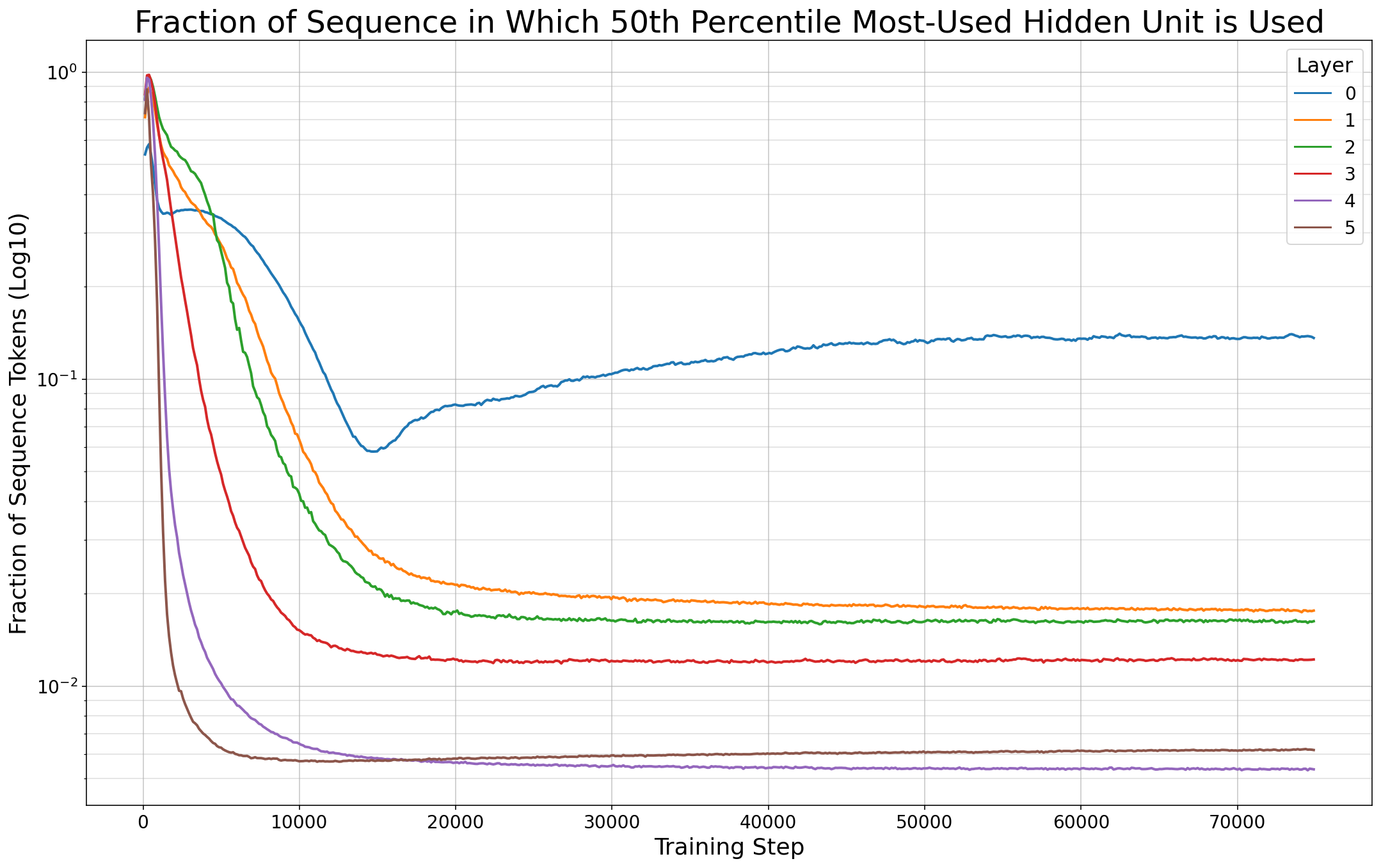}
  \caption{Use Frequency per Sequence of 50th Percentile Most-Used Hidden Unit (Log Scale)}
  \label{token_over_training:percentile}
\end{subfigure}
\caption{These plots show the early evolution of (\ref{token_over_training:token}) the number of dimensions activated per token, and (\ref{token_over_training:percentile}) the fraction of the sequence in which the 50th percentile most-used hidden unit in a sequence occurs.}
\label{token_over_training}
\end{figure}

Beyond converging to different levels of sparsity, there seem to be meaningful patterns of when in training each layer becomes sparse. Figure \ref{token_over_training} shows two metrics early in training: the fraction of hidden units active per token (again, simply the inverse of a traditional sparsity metric) and the median fraction of tokens within a sequence that use a hidden unit (excluding hidden units that \textit{never} activate in the sequence). These metrics are explained in more detail in \ref{metrics_discussion}. Both of these values drop early in training, and in particular both drop first in the final layer of the network, then the penultimate layer, and then the lower layers in nearly-monotonic succession from highest to lowest. 


Another important note from Figure \ref{token_over_training:percentile} is that unlike the other layers, Layer 0 reverses its trend and gradually uses each of its hidden units more frequently over the course of a sequence, where other layers are moving in the opposite direction. This is especially striking since, in Figure \ref{batch_dims}, we can see that during this period of training the total number of hidden units being used per batch drops precipitously for Layer 0, meaning that it is using fewer hidden units \textit{overall} but using the ones it does use \textit{more frequently}. We do see some tendency for other layers to undergo a drop in batch dimension starting at a similar point (in particular layers 1 and 2), but the degree is much less, and tends to partially recover after an initial dip. 

\section{Discussion}

\subsection{"Dead Neurons" and Unused Hidden Units}
\subsubsection{When Can We Declare a Neuron Dead?}
If a hidden unit (or, ``neuron'', to echo historical literature, is never nonzero over the 8192 tokens in a batch, is it ``dead'', in the conventional sense of that term? This is an interesting question because the term has historically encompassed two somewhat vaguely-defined properties: that it takes a value of 0 everywhere (or at least everywhere empirically sampled), and thus doesn't provide information to the model, and that it doesn't turn back on because it is on the wrong side of the ReLU. 

Based on the results in this paper, we believe we can usefully disentangle these concepts, and argue that, while is is possible for hidden unit that are empirically dead (i.e. never used in a large batch) to turn back on, there is also a point early in training where, if a hidden unit isn't being used, it has effectively been dropped by the model, in the sense that further training will not use it. 

We believe this on the strength of a test wherein, at 50,000 steps into training (out of a full training time of 1M steps), we calculate a mask of the hidden units by the current batch. This point was chosen due to it seeming in \ref{batch_dims} like a point at which all layers had stabilized how many hidden units they used per batch.  We then apply that mask as a zero-one masking on all future training steps, such that any hidden unit not active in the batch where the mask was calculated is constrained to be zero on future steps. We observed \textit{no statistical difference} in the accuracy achieved over the course of training between runs that did or did not have this masking applied. 

Concretely, by this point in training  naturally turned off about 30\% (calculated in aggregate over all layers) of its hidden units, and that we were able to perform the final 90\% of a training run with all of those hidden units masked, once the model had turned them off within a batch. As a check, we attempted to randomly select the same fraction of hidden units, and found performance collapse. This result suggests that, if you have independent reason to be training a ReLU model, it might be possible to exploit this emergent batch-level sparsity by reshaping your model to exclude unused dimensions from the latter ~90\% of training. We note that it may be possible to find ways to reduce sparsity and regain the value of that capacity, but we leave that to future work. 

Returning to the question at the beginning of this section: a hidden unit being unused over a batch during the very early stages of training shouldn't mark it as definitively dead, but by slightly later in training it is strong evidence that hidden unit could be removed without cost.

\subsection{How do Neurons Die?}
The accepted understanding of how neurons become ``dead'', or drop out of active model use, is that the behavior emerges randomly or accidentally, due to outlying batches that contribute a highly negative gradient impulse and knock the dimension into a regime where it stops activating. We believe like these results add complexity to and fundamentally challenge that mental model. \\

First, as mentioned in the previous section, a neuron or hidden unit being empirically dead within a batch does not mean that it cannot be turned on by the training process. It seems, then, strange to refer to these hidden units as dead, if that death can be reversed. Intuitively, this aligns with the fact that even the earliest MLP in a Transformer sits on top of an embedding layer and an attention mechanism, both of which can change the input distribution a dense layer experiences, and thus shift the distribution of its outputs even if the weights of the vector within the dense layer itself do not change. \\ 

Additionally, if ``neuron death'' were a random or accidental process, one would expect it to happen with similar frequency over the course of training, or at least to happen to a degree that scales to learning rate. Instead, in Figure \ref{batch_dims} we see very few dimensions turned off for the first 8000 steps of training - which for this model encompasses the period of highest learning rate before cosine decay begins - and then a dramatic increase after that. We believe these two facts are suggestive that sparsity is being driven by the model's construction of certain feature spaces to operate in, rather than simply happening accidentally and pathologically.

\subsection{Feature Specificity}
We hypothesize that some of the sparsity behavior observed here can be explained by something we here call \textit{Feature Specificity}. To explain the concept, imagine two ways features could be constructed. These are both extremes, and we don't expect either represents the internal behavior of real models, but we find them helpful models to illustrate a spectrum of possibilities. 

\begin{enumerate}
    \item \textbf{Continuous Shared Subspace} In this paradigm, all tokens have informative (nonzero) values along all dimensions, and information is communicated through the continuous differences in the values along each dimension.
    
    \item \textbf{Binary Representation} In this paradigm, there are many features, features take on binary values, and the information content of a token is represented by the set of hidden units which are on. In this world, you would expect the number of dimensions that are on (or 1) to differ dramatically from token to token, and would expect most dimensions to only be used in a small fraction of tokens, because if all dimensions were switched on all the time, there would be very little information to differentiate tokens. Intuitively, the lower information content of a 0/1 signal compared to a fully continuous one means that the full set of dimensions available to switch on or off would need to be larger in this paradigm to capture a similar amount of information. 
    
\end{enumerate}

Examining tables \ref{used_at_convergence} and \ref{percentiles_table}, as you rise higher in the network, features appear to shift from being closer to the continuous model to being closer to the binary model. Layer 0 is the most extreme case -- it converges to active use of a substantially smaller set of hidden units (\ref{used_at_convergence}), but without a commensurate drop in number of hidden units used per token. We can also see in Table \ref{percentiles_table} that hidden units in layer 0 are used much more frequently over a sequence: the 65th percentile hidden unit activates for at least 31\% of tokens in a sequence. The next highest layer on this metric, layer 2, activates for only of only 3.2\% of tokens, nearly 10x smaller. 

On the other extreme, we see layers 4 and 5 behaving in ways that more closely correspond with the binary feature model. They have the highest number of hidden units in active use per-batch (28.2K and 31.3K respectively), but use comparatively few for each token token. Looking at \ref{percentiles_table}, we see that the hidden units in these final two layers activate only very infrequently in the course of a sequence; the 90th percentile hidden units only activate 3.7\% and 6.9\% of tokens in a sequence respectively. To make this more concrete, for layer 4, out of dimensions that are ever used in a batch, 90\% of those dimensions only have nonzero values for 20 tokens or fewer out of a 512-length sequence. 

\subsection{How Does Sparsity Relate to Model-Usable Capacity?}

\begin{table}[]
\centering
\begin{tabular}{ccccc}
Layer & \begin{tabular}[c]{@{}l@{}}Hidden Units Used \\ Round 1\end{tabular} & \begin{tabular}[c]{@{}l@{}}Hidden Units Used \\ Round 2\end{tabular} & \begin{tabular}[c]{@{}l@{}}Percent Used \\ Round 1\end{tabular} & \begin{tabular}[c]{@{}l@{}}Percent Used \\ Round 2\end{tabular} \\
0     & 7455                                                               & 1364                                                               & 22.7                                                            & 18.2                                                            \\
1     & 24807                                                              & 18529                                                              & 75.7                                                            & 74.6                                                            \\
2     & 22743                                                              & 16079                                                              & 69.4                                                            & 70.6                                                            \\
3     & 24478                                                              & 19455                                                              & 74.7                                                            & 79.4                                                            \\
4     & 27926                                                              & 24171                                                              & 85.2                                                            & 86.5                                                            \\
5     & 32200                                                              & 31730                                                              & 98.2                                                            & 98.5                                                           
\end{tabular}
\caption{This table shows the results of an experiment where, in the first training round, all layers had access to a maximum hidden dim of 32768, of which the values in ``Round 1`` were used at convergence. In Round 2, the maximum hidden dim for each layer was set to the value from the ``Round 1`` column, and ``Round 2`` shows how many of those available dimensions were used.}
\label{table:4}
\end{table}

On a very simplistic level, all of this batch-level sparsity seems like a waste: capacity that the model could be making productive use of that it appears to be ignoring. In one framing, the number of dimensions the model converges to for each layer could be a sort of revealed preference of capacity - perhaps for this width or depth or dataset, the capacity needs of some layers are bottlenecked by others. \\

To investigate this as a possible framing, observed the number of hidden units used at convergence in a standard experimental setting. We then started a new model run wherein each layer was given that number of hidden units as its maximum MLP hidden dim. If the previously observed sparsity was indicative of the capacity the model could effectively utilize, one would expect to see very little sparsity emerge in this second experiment, since the model would need to use every dimension available to it to reach the same capacity. Instead, we see the model trained in Round 2 uses a similar per-layer fraction of its available capacity to what we saw in Round 1, when every layer started with 32768 dimensions of available capacity. This is despite the fact that the model at round 2 has \textit{substantially worse performance}, indicating this capacity reduction is costly. \\ 

From this we hypothesize that each layer has a certain fractional subset of initialization space that it is able to effectively use for the features that layer preferentially learns, and when you give the network a lower total hidden dim, you proportionally reduce the number of dimensions that fit this criteria. We weren't able to more fully explore this hypothesis in this work, but believe it could be an important area for future study.

\section{Sensitivity to Hyperparameters}
\label{hyperparameter_tests}
Thus far we have focused single case study model to allow for more in-depth examination. Here we seek to understand how these sparsity patterns generalize across different hyperparameter settings. For all experiments discussed here, plots showing behavior over the course of training can be found in the appendix.

\subsection{Sensitivity to Learning Rate}
\begin{table}[]
\centering
\begin{tabular}{ccccc}
\textbf{Layer} & \textbf{\begin{tabular}[c]{@{}l@{}}Batch Use (\%)\\ (LR=3e-3)\end{tabular}} & \textbf{\begin{tabular}[c]{@{}l@{}}Batch Use (\%)\\ (LR=2e-3)\end{tabular}} & \textbf{\begin{tabular}[c]{@{}l@{}}Batch Use (\%)\\ (LR=1e-3)\end{tabular}} & \textbf{\begin{tabular}[c]{@{}l@{}}Batch Use (\%)\\ (LR=5e-3)\end{tabular}} \\
0              & 13.3                                                                    & 24.8                                                                    & 37.4                                                                   & 19.9                                                                    \\
1              & 79.3                                                                   & 83.2                                                                   & 93.4                                                                   & 75.6                                                                   \\
2              & 71.0                                                                   & 63.5                                                                   & 55.8                                                                   & 76.4                                                                   \\
3              & 73.0                                                                   & 71.7                                                                   & 78.4                                                                   & 79.6                                                                   \\
4              & 86.1                                                                   & 92.1                                                                   & 99.1                                                                   & 80.0                                                                   \\
5              & 95.6                                                                   & 98.0                                                                   & 99.0                                                                   & 88.1                                                                   \\
\textbf{Total}          & \textbf{69.7}                                                                 & \textbf{72.2}                                                              & \textbf{77.2}                                                                  & \textbf{69.9}                                                                 
\end{tabular}
\caption{The fraction of hidden units used per-batch for different tested learning rates (the first column represents the standard experimental setting).}
\label{batch_dims_lr}
\end{table}

Our optimizer is structured to use a warm-up period of 5000 steps, up to a peak learning rate, and then cosine decay thereafter. Table \ref{batch_dims_lr} examines how the converged Per-Batch Hidden Unit Use of a 6-layer model varies based on peak learning rate. Note that the standard experimental setting of 3e-3 has the highest performance, since it was selected via hyperparameter search. Some things are clearly consistent over these different learning rates: in all cases layer 0 substantially fewer dimensions over the course of a batch compared to any other layer, and the two highest layers of the network still use the most dimensions, with the final layer continuing to use the most dimensions in all but one case where layer 4 overtakes it by a minute margin. Overall, aggregate sparsity levels, and particularly the extreme batch-level sparsity in layer 0, goes down as peak learning rate decreases. 

\subsection{Sensitivity to Hidden Dimension}

\begin{figure}
\begin{subfigure}{.5\textwidth}
  \centering
\includegraphics[width=\linewidth]{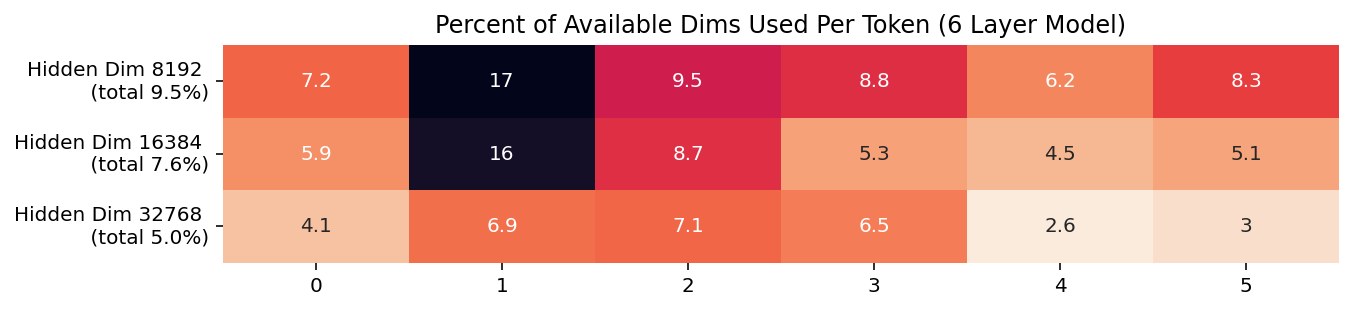}
  \caption{Dimensions Per Token by Hidden Dim}
  \label{hidd_dim_fig:token}
\end{subfigure}%
\begin{subfigure}{.5\textwidth}
  \centering
  \includegraphics[width=\linewidth]{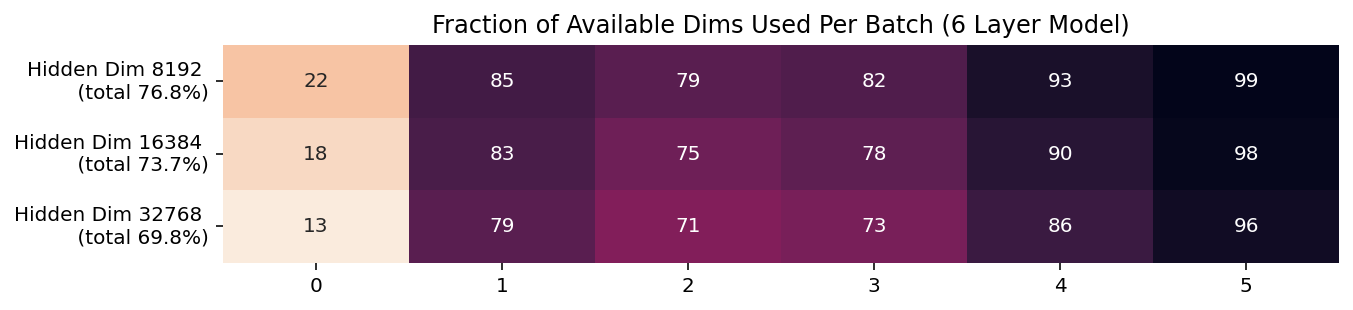}
  \caption{Dimension Use Per Batch by Hidden Dim}
  \label{hidd_dim_fig:batch}
\end{subfigure}
\caption{These plots show the fraction of dimensions used per-batch or per-token, for models of hidden dimension 8192, 16384, 32768. }
\label{hidd_dim_fig}
\end{figure}


In \cite{li2023lazy} the authors showed that wider models had generally higher levels of sparsity, and we see see the same directional result here in \ref{hidd_dim_fig}, both for hidden units used per token (the measurement closest to the one used in that paper) and for hidden units used per batch. This would be consistent with the idea that there is a minumum necessary absolute number of hidden units for good performance (and thus a higher fraction of a narrower model), but that after that the marginal hidden unit declines in value. 

\subsection{Sensitivity to Model Depth}


\begin{figure}
  \centering
  \includegraphics[width=.7\linewidth]{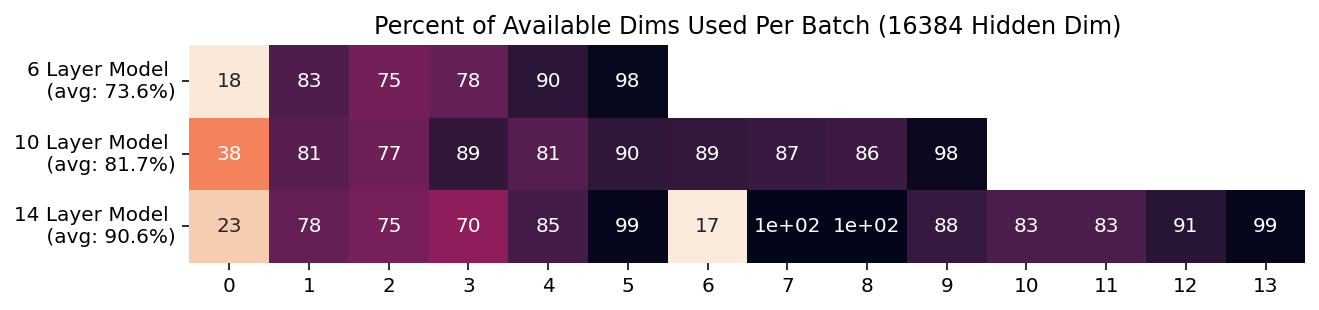}
  \caption{The fraction of available hidden units used per batch for a 6, 10 and 14 layer model with a hidden dimension of 16384.}
  \label{depth_fig}
\end{figure}


To explore the sensitivity of our observations to model depth, we trained a 6, 10, and 14 layer model with a hidden dimension of 16384 (so that all models could fit on available hardware) In \ref{depth_fig} we can see that, when it comes to depth rather than width, we do not get the same directional result as above: there isn't a clear trend in hidden units used per token, but hidden units per batch tend to increase in a deeper model. One explanation that could be at play here is that as we deepen the network we create new layers that are more similar in their behavior to the later layers of a six layer model. That does seem broadly supported by the results here, though we seem to also see some cases (layer 6, in the 14 layer model) which more closely follow the pattern of layer 0 in using a very small fraction of its hidden units per batch, suggesting that as models get deeper there may be value in additional instances of multiple of the layer archetypes so far discussed, not just those of the later layers. 


\section{Limitations}
The primary limitations of this work come from the fact that we chose to focus exclusively on small models and ReLU activation functions, both of which pull these conclusions further away from immediate applicability to modern foundation models. The former choice was constrained by the amount of metric iteration and the number of ablations this work entailed, which would have been infeasible to justify with large models. The latter was motivated by the simple fact that ReLUs induce the kind of hard sparsity that is easier to study. Based on the results in \citet{mirzadeh2023relu}, where the authors are able to sparsify models not trained with ReLU, we are hopeful that there are underlying similarities in the features being learned between these model types, but which may be easier to bring into focus in a fully sparse setting. We hope that future work along these lines can identify proposed changes to initialization functions or other aspects of model structure that will allow for more effective use of capacity across both ReLU and more competitive activation functions. 

\section{Related Work}

\textbf{Examination of Empirical Sparsity} \citet{li2023lazy} empirically examines the layer-by-layer token sparsity dynamics of trained encoder-decoder models, focusing on T5. On a high level, we see low-single-digit fraction of hidden dimensions used per token, which is consistent with their findings, though the fact of choosing to focus on decoder-only transformers and smaller networks means we wouldn't necessarily expect to be able to replicate their work directly. Intrigued by the conclusions of their work, we expand to look not only at the sparsity properties of a fully trained model, but the way sparsity evolves over training. We also widen the scope of this previous work by taking an integrated view of sparsity - both at the token level, and across sequences and batches.

\citet{voita2023neurons} also explores layer-dependent sparsity, and finds that many neurons "die", in the sense of never being used over a broad set of data, predominantly in early layers, but only look at this behavior at the end of training, while we explore the evolution of sparsity over the course of training.

\citet{lu2020dyingreluinit} similarly finds that some hidden units consistently have a value of zero post-RELU at the beginning of training, and suggest improvements to initialization to address this. However, their derivations on a simple feed-forward network appear to make assumptions about the inputs to each successive layer that it is unclear would apply to the structure of Transformer blocks with residual connections in between. 

\citet{waleffe2020principal} argues through use of PCA that, in the convolutional image models they study, activations only actually use a subspace of smaller rank than the full available hidden dim, and that this can be used to motivate parameter reductions early in training. This is consistent with our findings in language transformers that many activation dimensions are almost never used from early on in training.

\textbf{Induced Sparsity} Outside of this, the largest cluster of similar work attempts to induce usable, structured sparsity beyond the amount that natively results from training. \citet{jaszczur2021sparse} argues that it's possible to both sparsify attention and to convert dense MLPs to use a top \(K\) operation, with minimal cost to performance. \citet{runwal2024peft} designs a fine-tuning loss that disincentivizes dense use of dimensions, though that work focuses on encoder-decoder models rather than decoder-only. \citet{szatkowski2024sadmoe} and \citet{zheng2024learn} operate in the setting of "MoEification" where a normally trained model is converted to function as a MoE, and each proposes different procedures for inducing activation sparsity in the underlying model to improve the results of this process.

Recent work by \citet{mirzadeh2023relu} shows that you can take a LLM trained with GELU or SiLU activation - which have no bias towards sparsity - and convert it to use ReLU activation with minimal performance loss through a short period of fine-tuning. Once the model is converted to ReLU, the authors find high degrees of activation sparsity. This is compelling because it suggests a degree of "latent sparsity" even in models trained with smooth activation functions, and, in our view, makes it more likely that the sparsity dynamics we find here may also correspond to similar dynamics in smoothly-activated networks, but whose effects are easier to legibly study in the ReLU context.

\section{Conclusion}
This work seeks to expand the default understanding of sparsity in transformers, from being understood as a generally shared and roughly uniform property across layers, to one that varies in its characteristics at different depths of a network in ways that strongly appear to be systematic. We show evidence demonstrating that the first and last layers of a network are effective inverses of one another in terms of the sparsity behaviors they exhibit, and suggest that this may be due to underlying differences in the structure of features learned at different layers, with layer 0 learning features more densely distributed across a smaller vector space, and later layers, especially the final layer, learning features that are closer to binary over a larger vector space. On a less scientific and more pragmatic front, we demonstrate that, if you have independent reason to be training a ReLU model, you might be able to prune 30+\% of the model's hidden dimension neurons for most of training without meaningful cost to accuracy, and with a potentially large savings on compute. \\

\medskip

\bibliography{references}

\appendix

\section{Appendix}
The appendices are organized as follows: 
\begin{enumerate}
    \item The first two sections focus on providing expanded or additional context to the core experimental setting. \ref{full_plots_core} showing full training curve plots for cases where only the beginning of training was shown in the main paper. The main body of the paper shows aggregated metrics for measures of how frequently hidden units are used within a sequence, and shows a plot for the median (50th percentile). \ref{higher_percentiles} expands on this, and shows plots for the 65th, 75th, and 90th percentile metrics over the course of training.
    \item The final three sections focus on providing training curve plots for the hyperparameter ablations discussed in section \ref{hyperparameter_tests}, where only aggregations were shown in the full paper. 
    \item The final section in the appendix, \ref{code_definitions}, gives pseudocode definitions for the core metrics discussed and used in the paper. 
\end{enumerate}

\subsection{Full Training Plots for Core Metrics}
\label{full_plots_core}
These plots show the full behavior over training of metrics for which a truncated training plot was shown in the full paper. The goal here is to demonstrate that interesting convergence behavior appears to be confined to the early stages of training, and justify the choice to focus on those in the main body of the paper. 
\begin{figure}[H]
    \centering
    \includegraphics[width=0.8\linewidth]{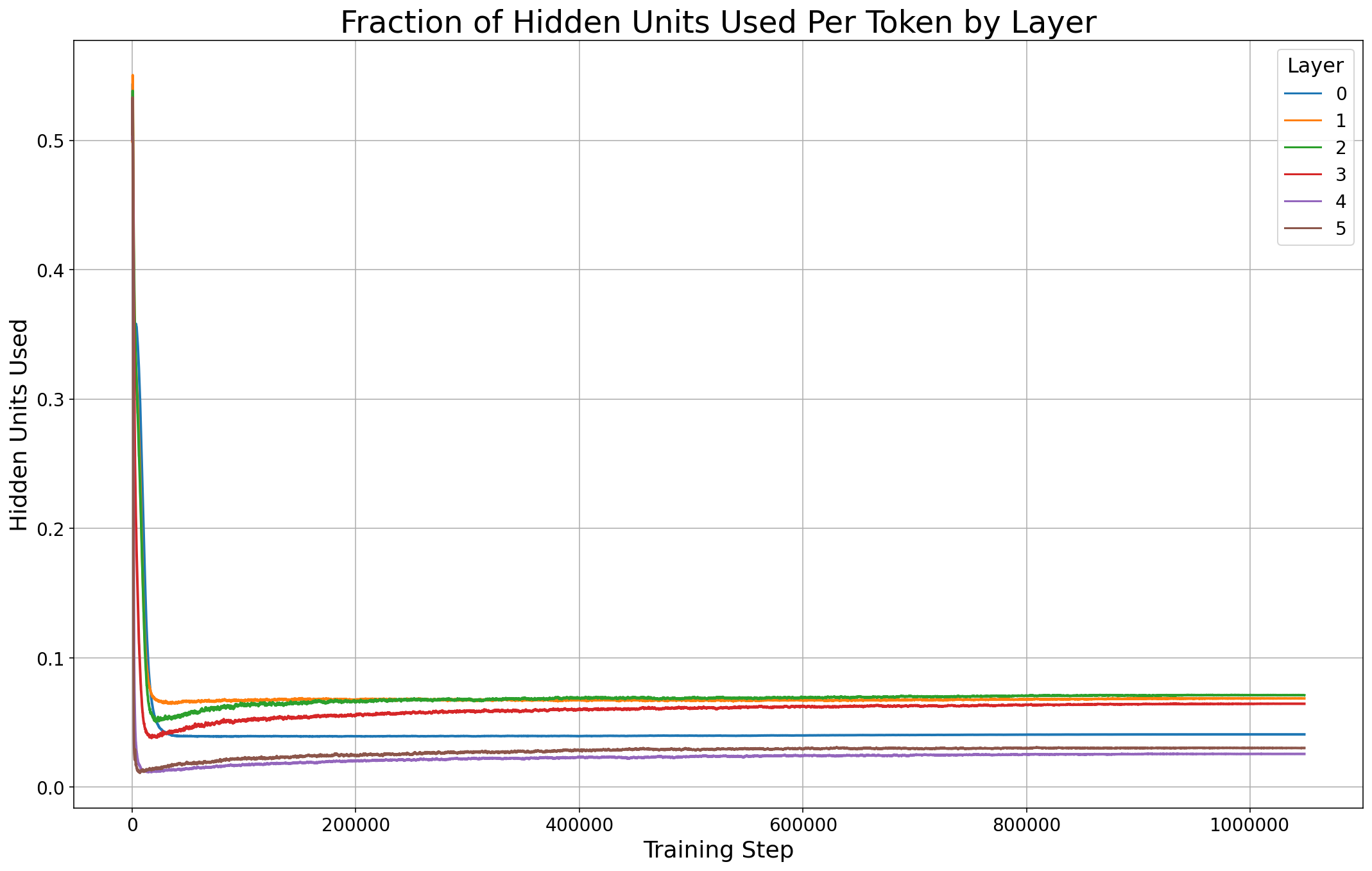}
    \caption{Hidden Units Used Per Token: Full Training Curve}
    \label{per_token_full}
\end{figure}
\begin{figure}[H]
    \centering
    \includegraphics[width=0.8\linewidth]{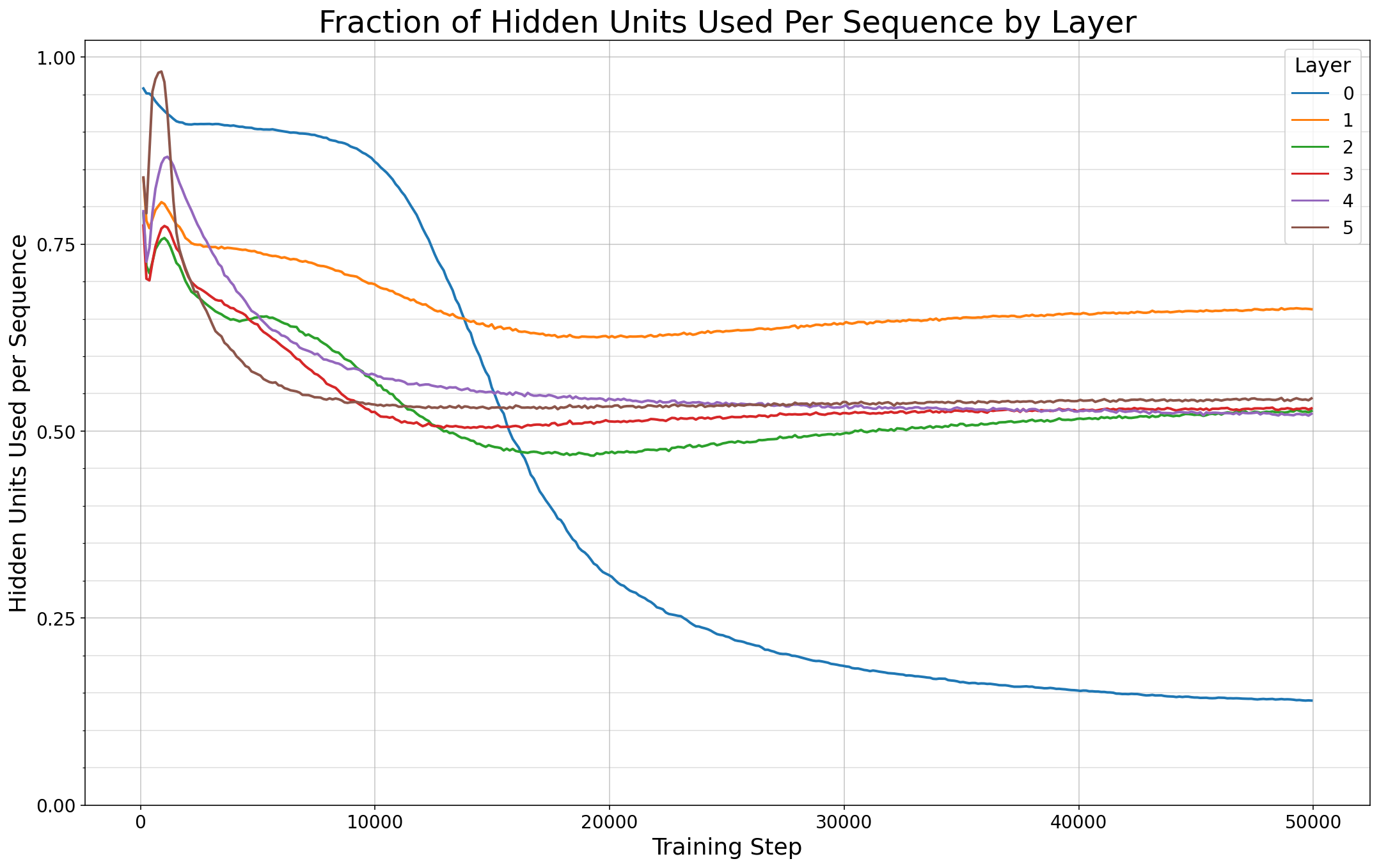}
    \caption{Hidden Units Used Per Sequence: Full Training Curve}
    \caption{The number of hidden units used total by a sequence, over the course of training.}
    \label{per_sequence_full}
\end{figure}
\begin{figure}[H]
    \centering
    \includegraphics[width=0.8\linewidth]{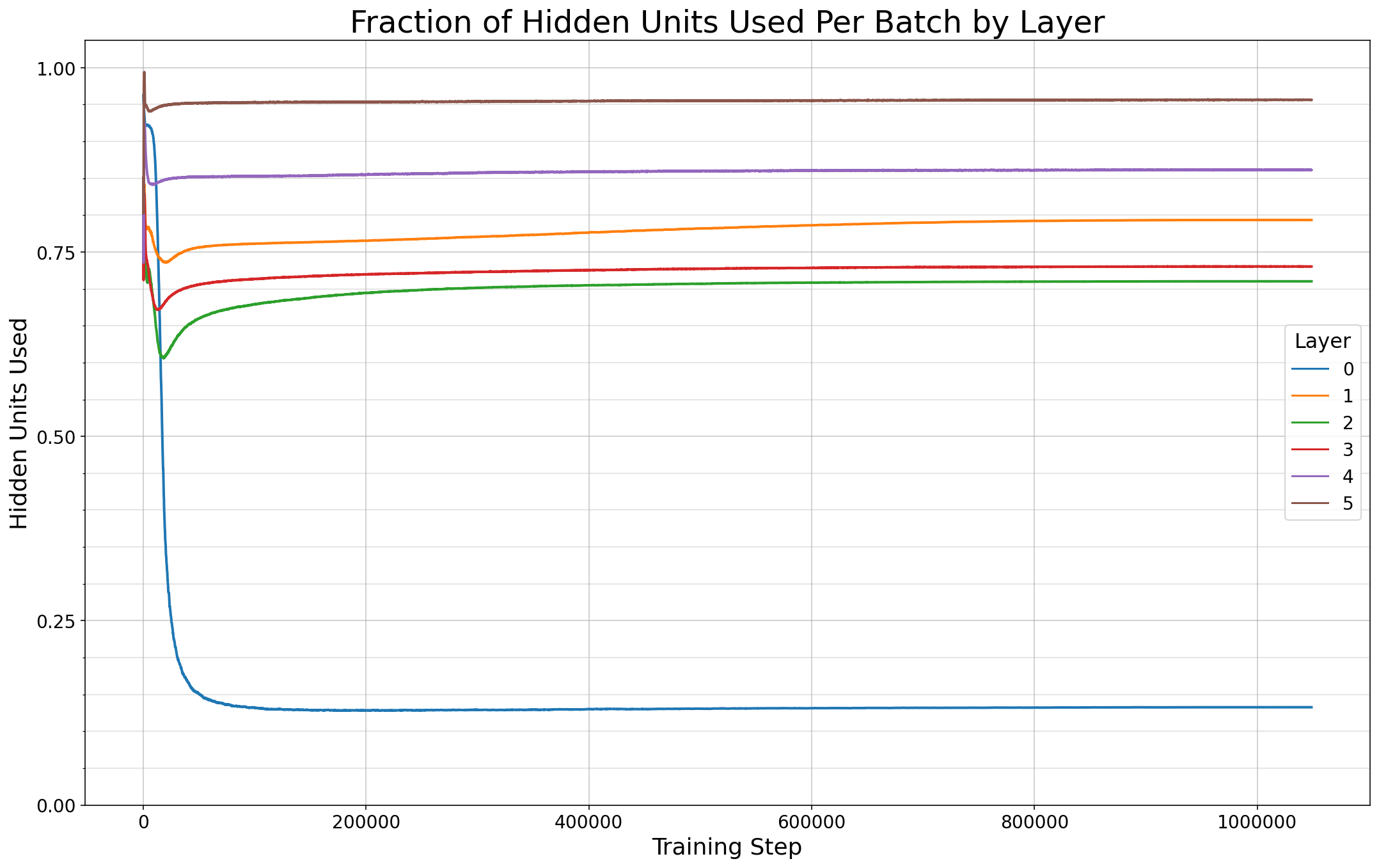}
    \caption{Hidden Units Used Per Batch: Full Training Curve}
    \caption{The number of hidden units used total by all sequences in a batch, over the course of training.}
    \label{per_batch_full}
\end{figure}

\subsection{Higher Percentile Plots of Hidden Dimension Use Frequency}
\label{higher_percentiles}

\begin{figure}[H]
\begin{subfigure}{.5\textwidth}
  \centering
\includegraphics[width=\linewidth]{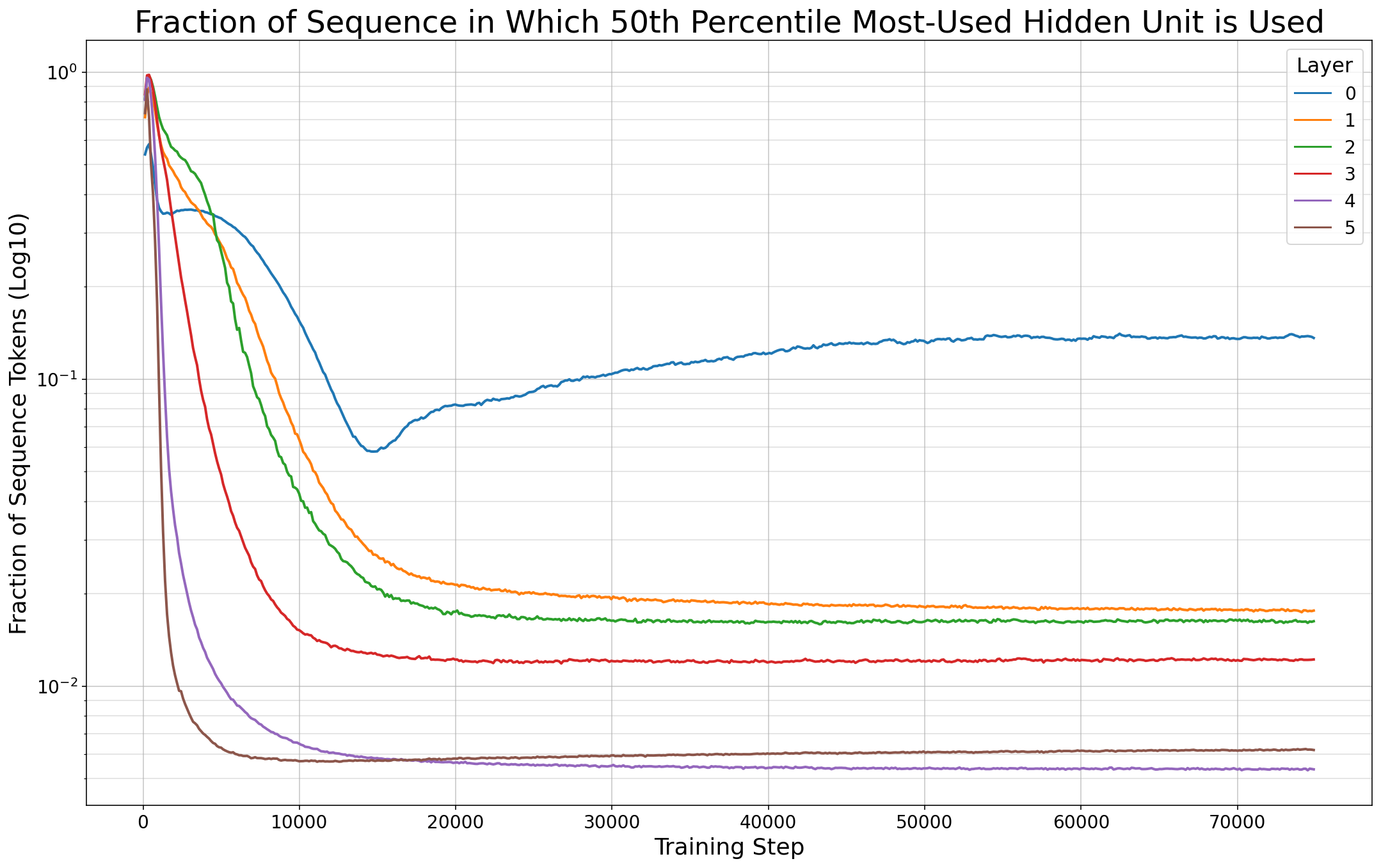}
  \caption{50th Percentile Hidden Unit Use: Short}
\end{subfigure}%
\begin{subfigure}{.5\textwidth}
  \centering
  \includegraphics[width=\linewidth]{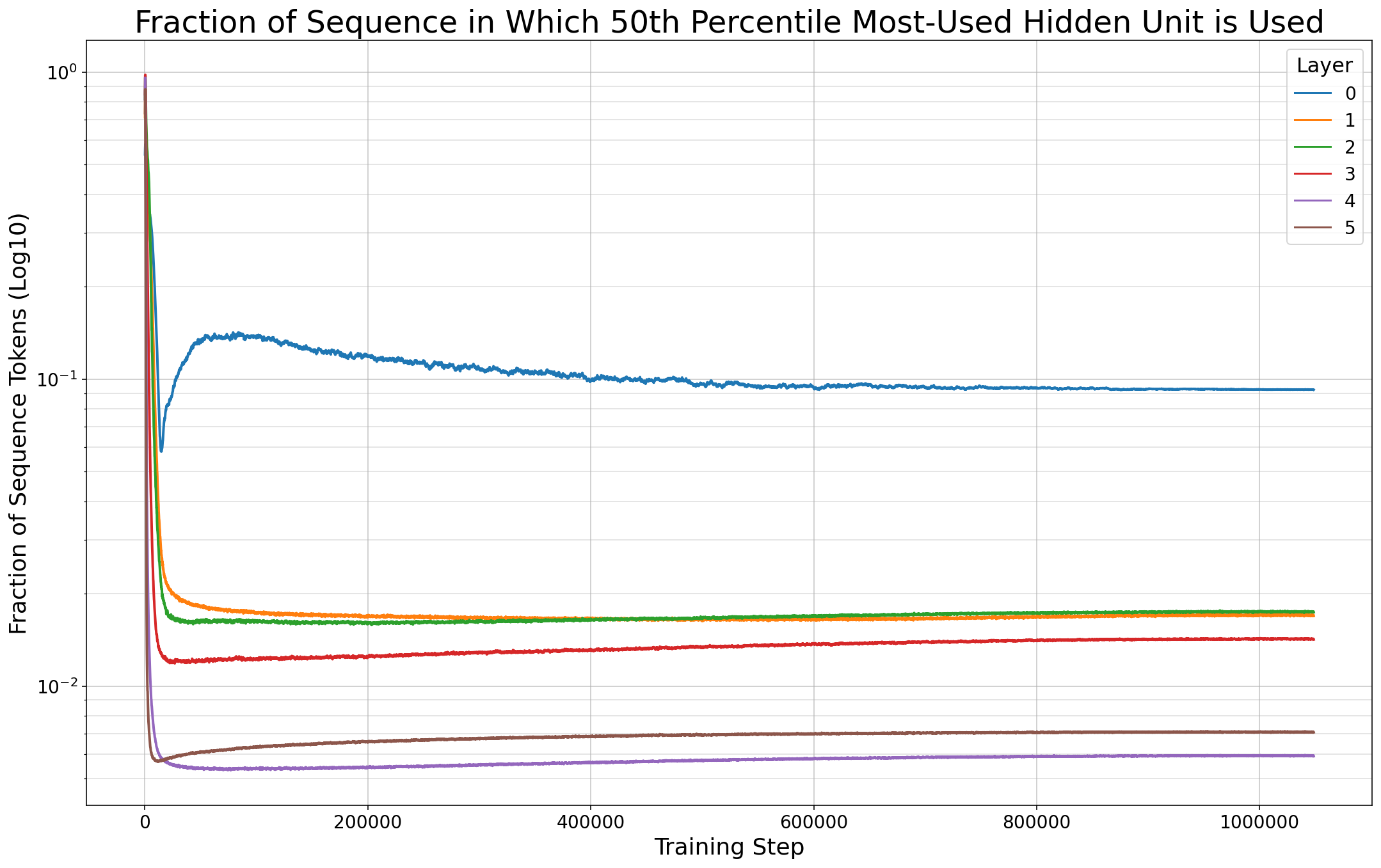}
  \caption{50th Percentile Hidden Unit Use: Full}
\end{subfigure}
\caption{These plots show the fraction of the sequence in which the 50th percentile most-used hidden unit in that sequence is used, at the beginning and over the full course of training.}
\label{50_perc}
\end{figure}

\begin{figure}[H]
\begin{subfigure}{.5\textwidth}
  \centering
\includegraphics[width=\linewidth]{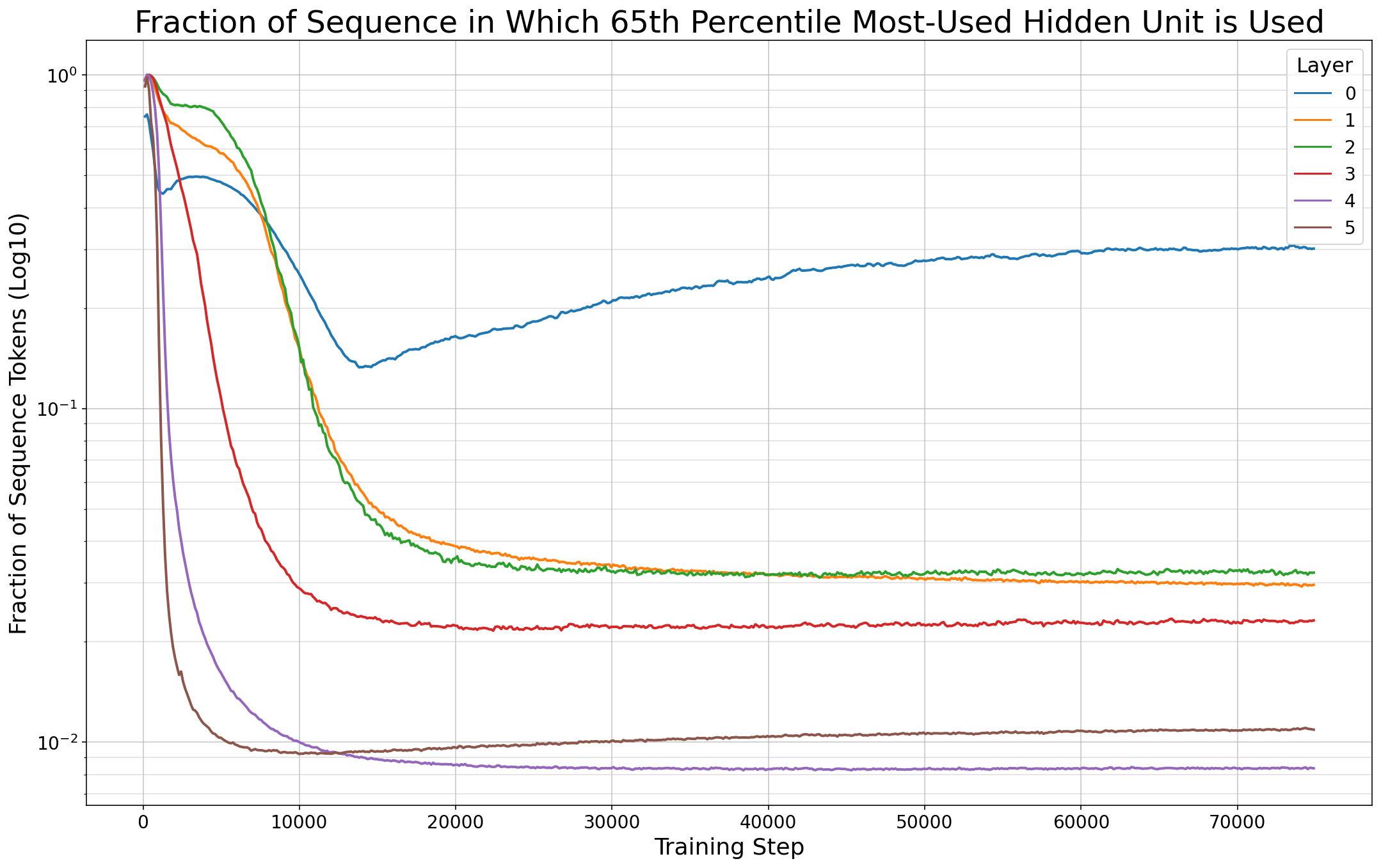}
  \caption{65th Percentile Hidden Unit Use: Short}
\end{subfigure}%
\begin{subfigure}{.5\textwidth}
  \centering
  \includegraphics[width=\linewidth]{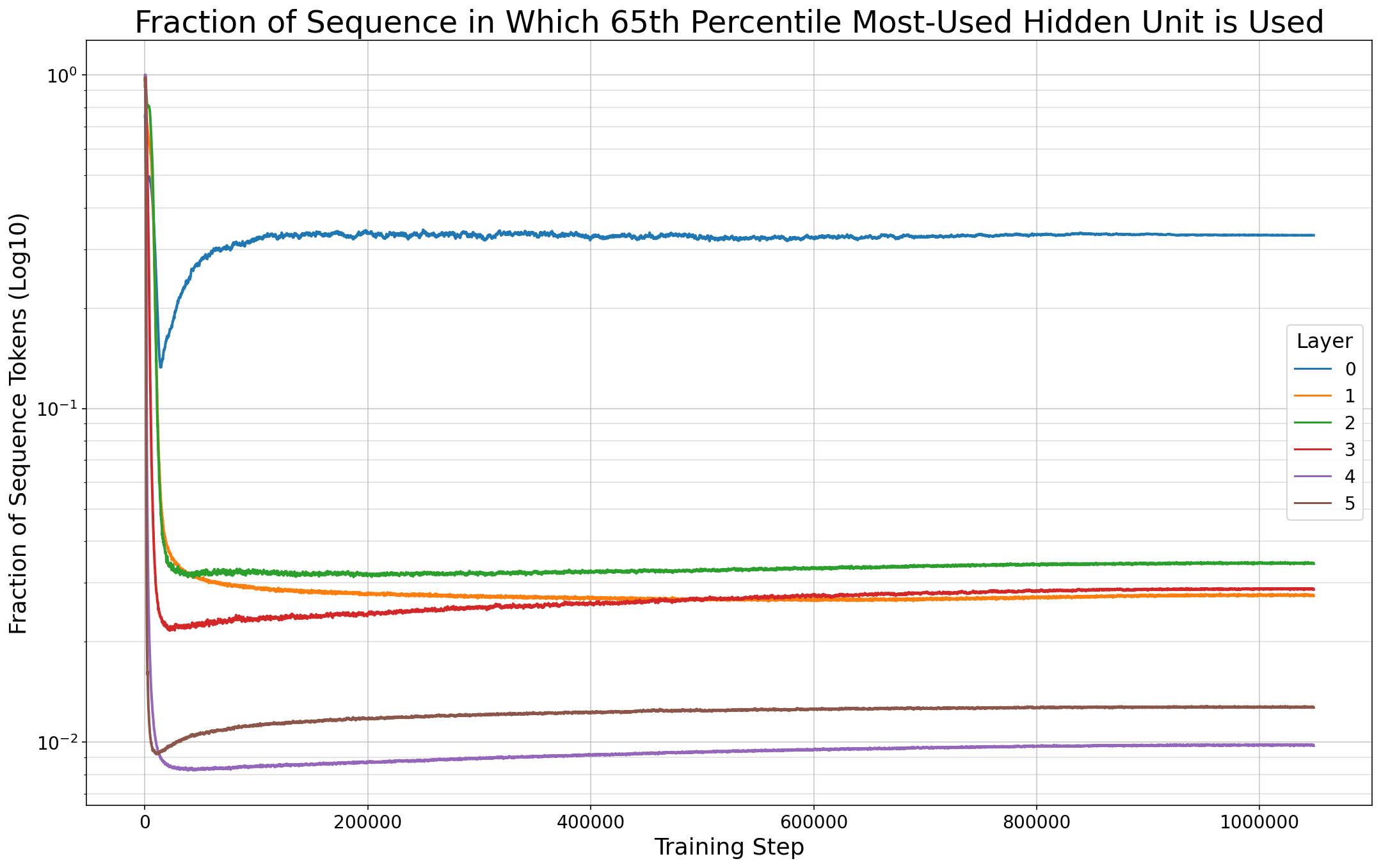}
  \caption{65th Percentile Hidden Unit Use: Full}
\end{subfigure}
\caption{These plots show the fraction of the sequence in which the 65th percentile most-used hidden unit in that sequence is used, at the beginning and over the full course of training.}
\label{65_perc}
\end{figure}

\begin{figure}[H]
\begin{subfigure}{.5\textwidth}
  \centering
\includegraphics[width=\linewidth]{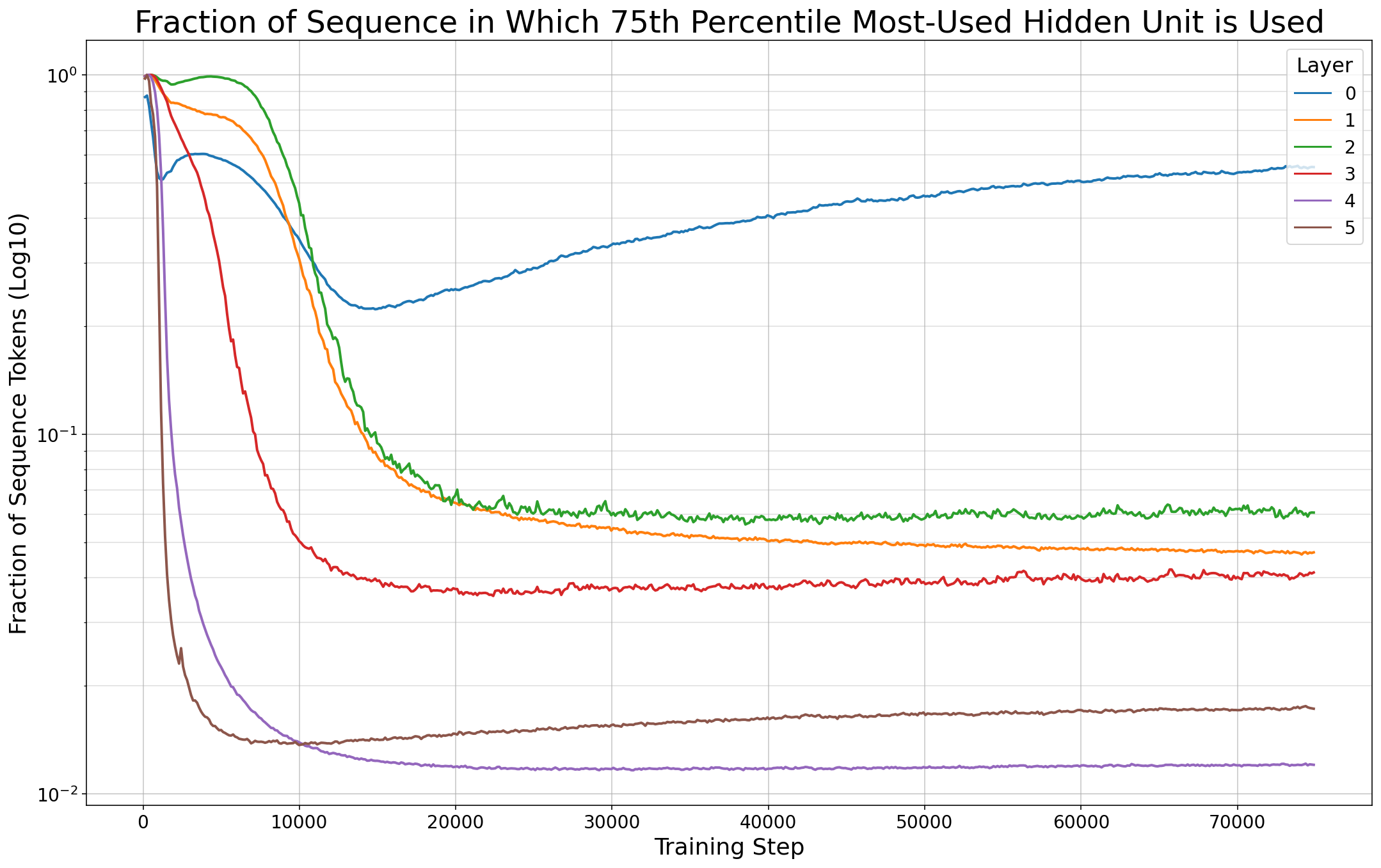}
  \caption{75th Percentile Hidden Unit Use: Short}
\end{subfigure}%
\begin{subfigure}{.5\textwidth}
  \centering
  \includegraphics[width=\linewidth]{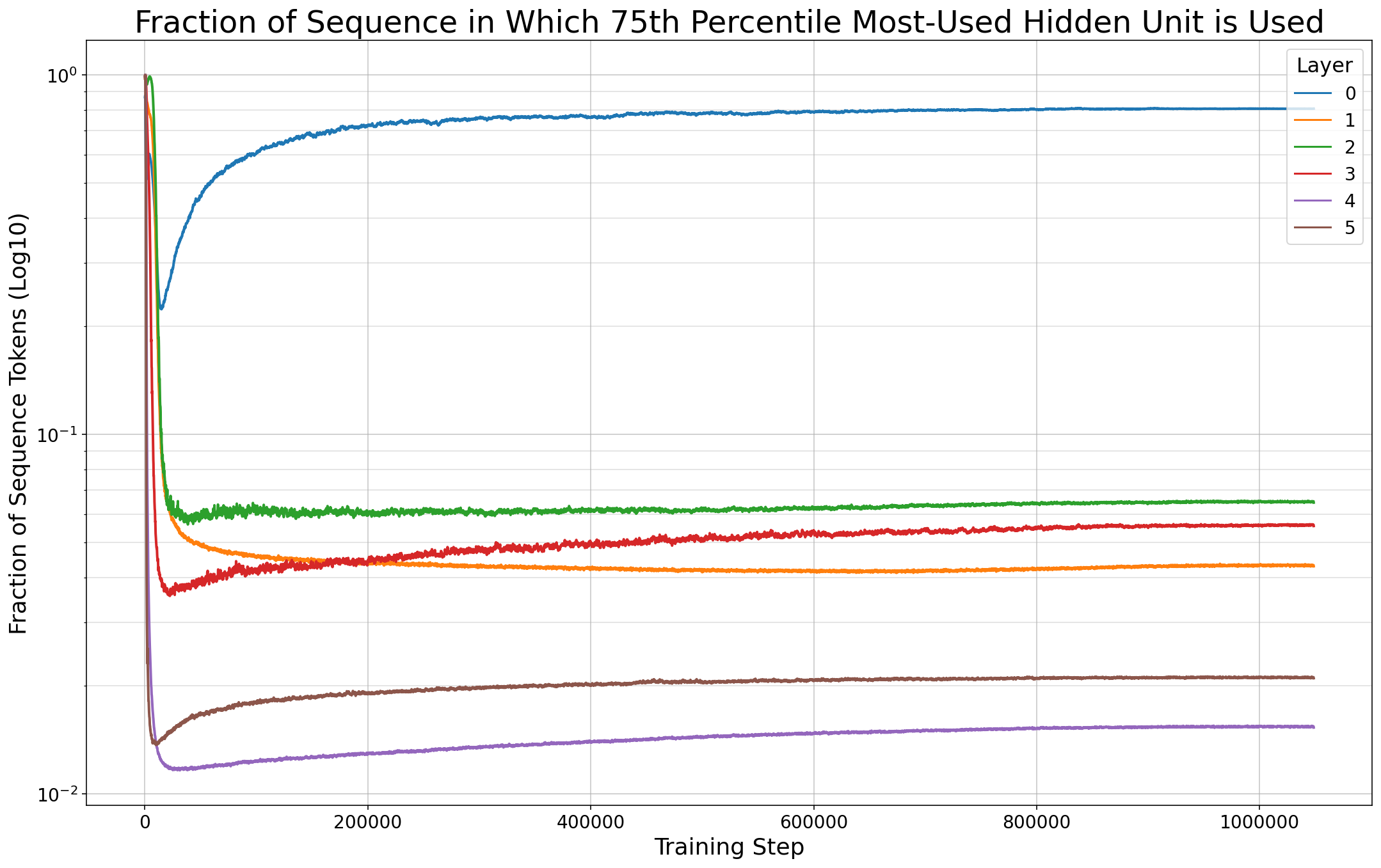}
  \caption{75th Percentile Hidden Unit Use: Full}
\end{subfigure}
\caption{These plots show the fraction of the sequence in which the 75th percentile most-used hidden unit in that sequence is used, at the beginning and over the full course of training.}
\label{75_perc}
\end{figure}
\begin{figure}[H]
    \begin{subfigure}{.5\textwidth}
        \centering
        \includegraphics[width=\linewidth]{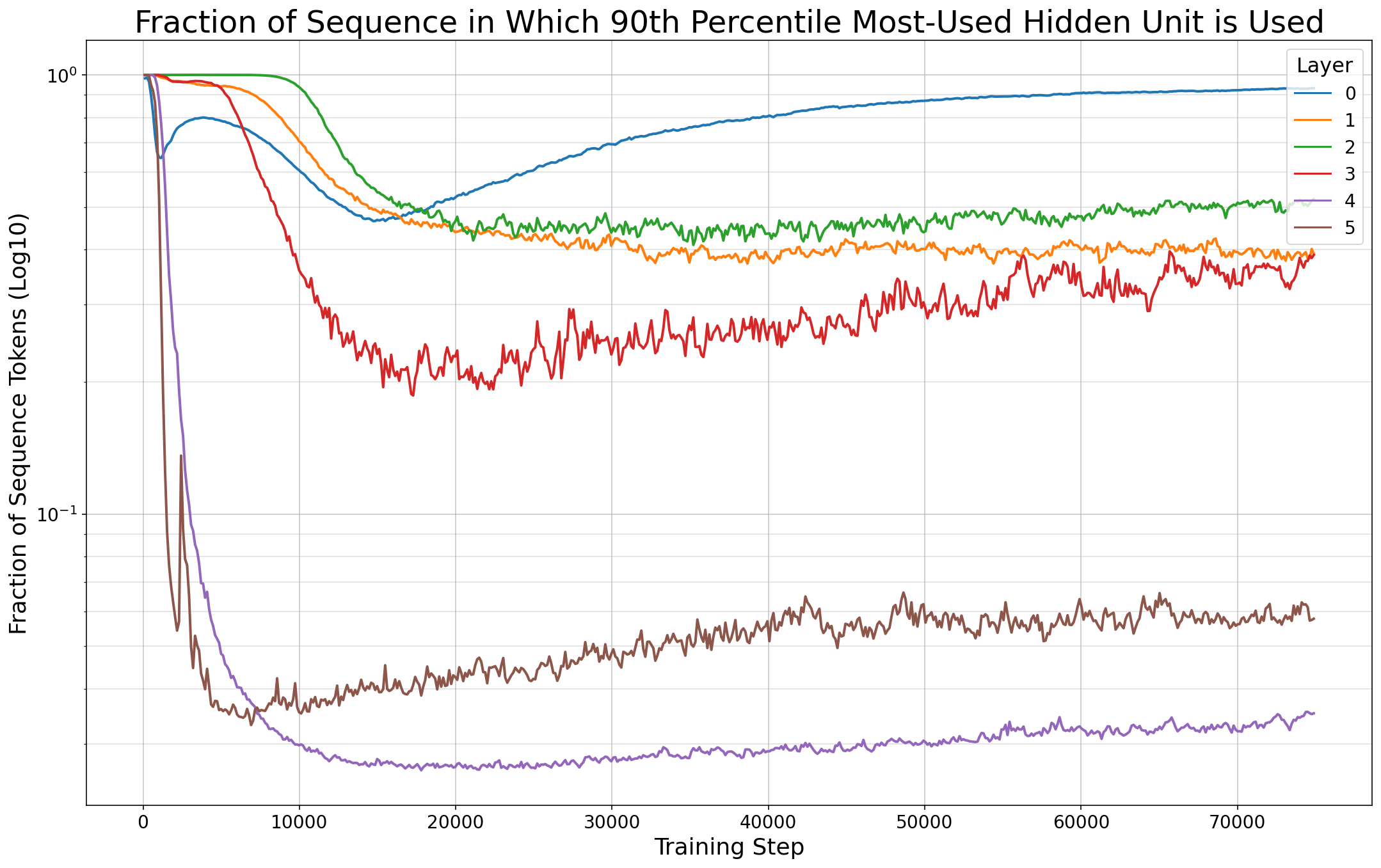}
        \caption{90th Percentile Hidden Unit Use: Short}
    \end{subfigure}%
    \begin{subfigure}{.5\textwidth}
      \centering
      \includegraphics[width=\linewidth]{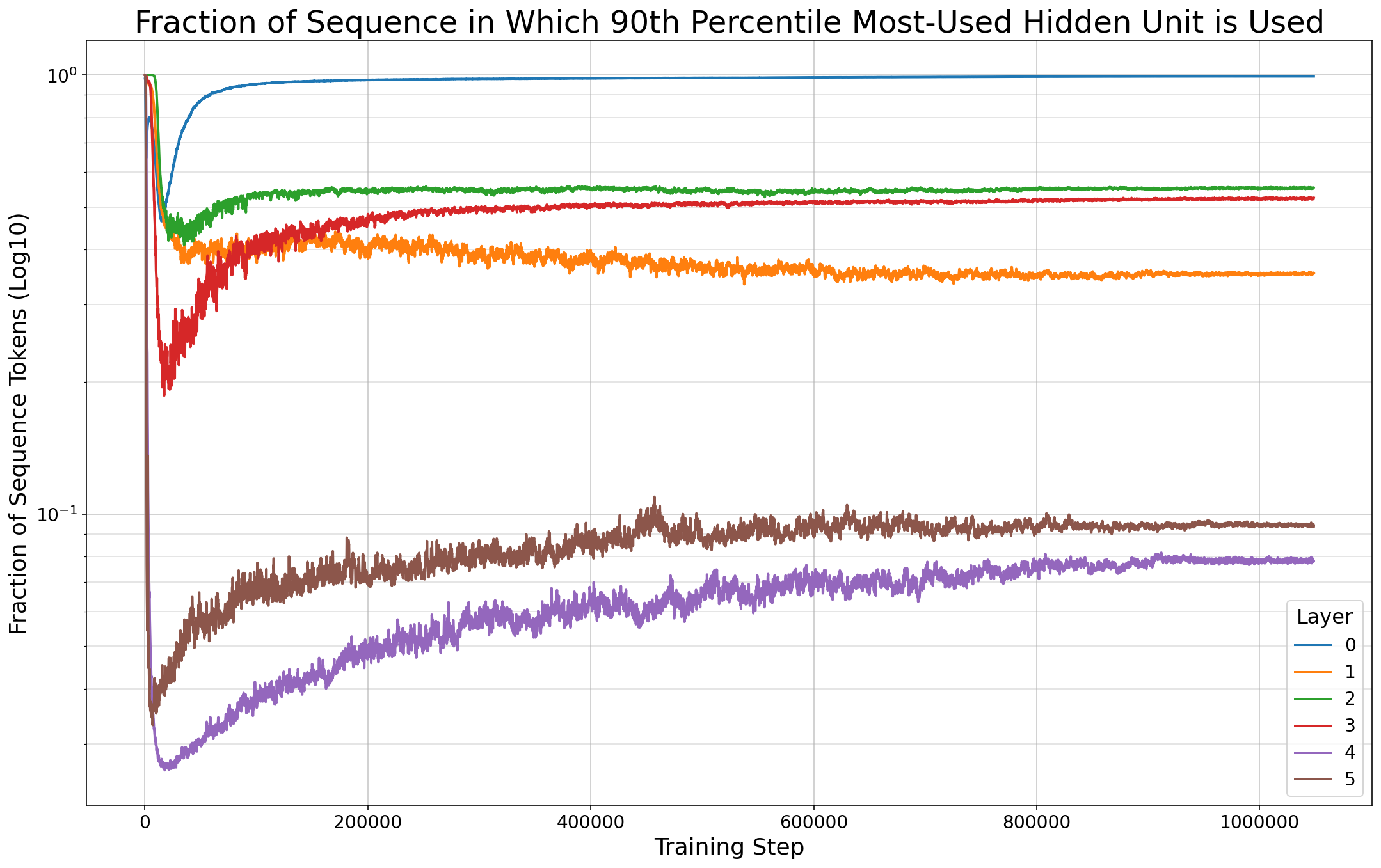}
      \caption{90th Percentile Hidden Unit Use: Full}
    \end{subfigure}
\caption{These plots show the fraction of the sequence in which the 90th percentile most-used hidden unit in that sequence is used, at the beginning and over the full course of training.}
\label{90_perc}
\end{figure}

\subsection{Sensitivity to Depth Plots}
\label{depth_sensitivity}
This section shows token, sequence, and batch hidden unit use for models of 6, 10, and 14 layers, all of which use a hidden unit size of 16384 so that even 14 layer can fit on available hardware. 

\subsubsection{6 Layer}
\begin{figure}[H]
    \centering
    \includegraphics[width=0.8\linewidth]{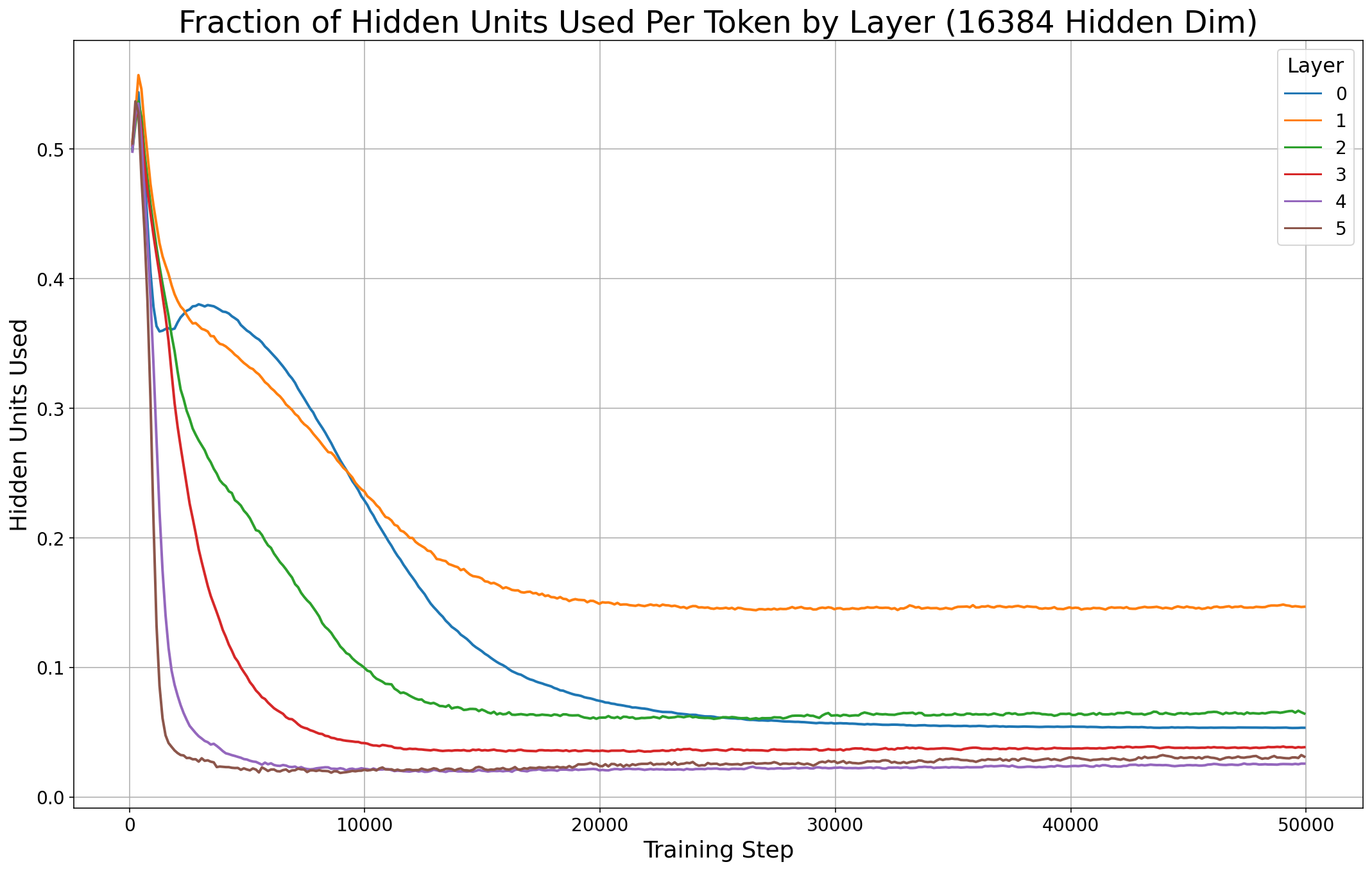}
    \caption{Fraction of available 16384 hidden units used per token in a 6 layer model.}
    \label{per_token_6l}
\end{figure}
\begin{figure}[H]
    \centering
    \includegraphics[width=0.8\linewidth]{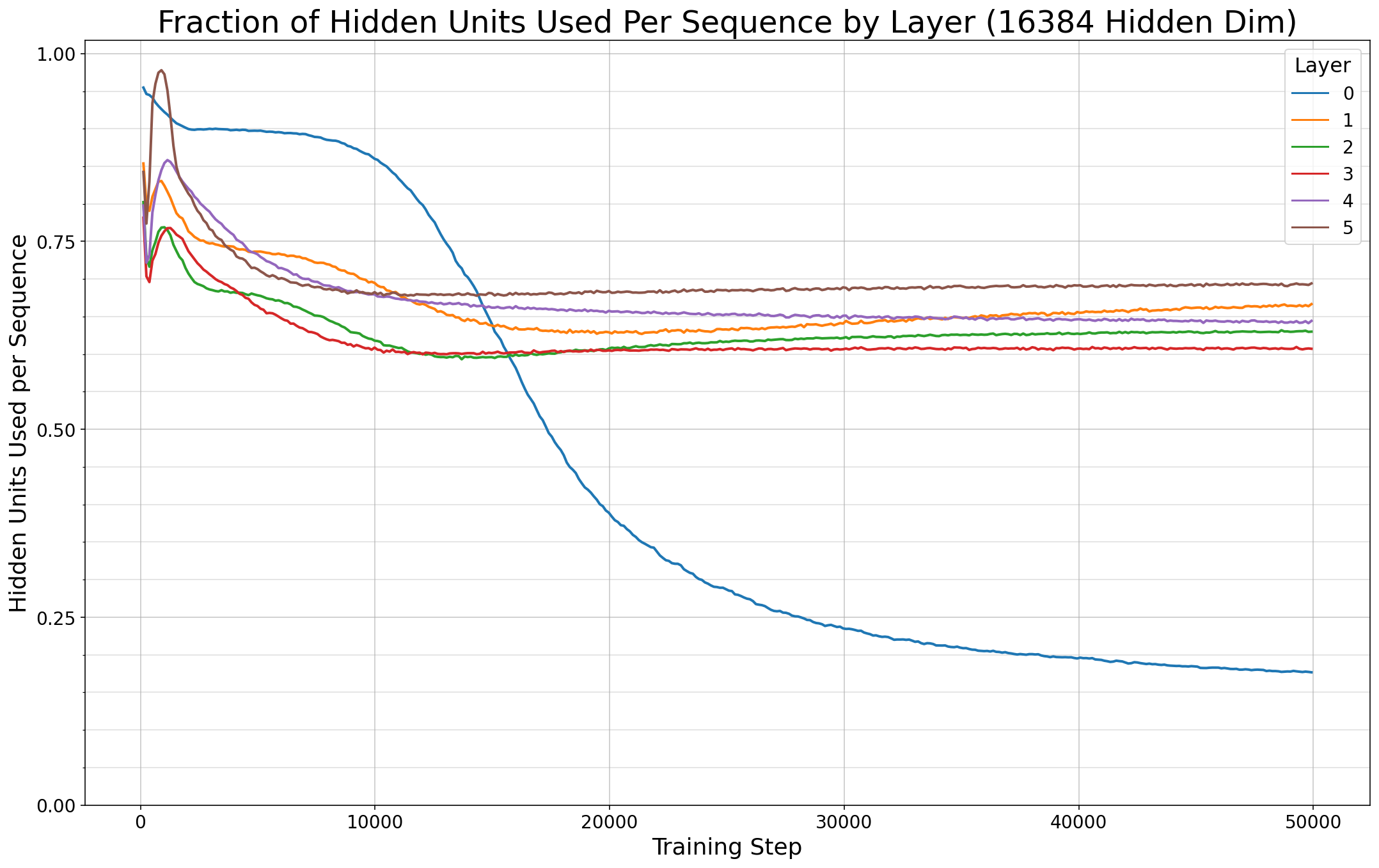}
    \caption{Fraction of available 16384 hidden units used per sequence in a 6 layer model.}
    \label{per_sequence_6l}
\end{figure}
\begin{figure}[H]
    \centering
    \includegraphics[width=0.8\linewidth]{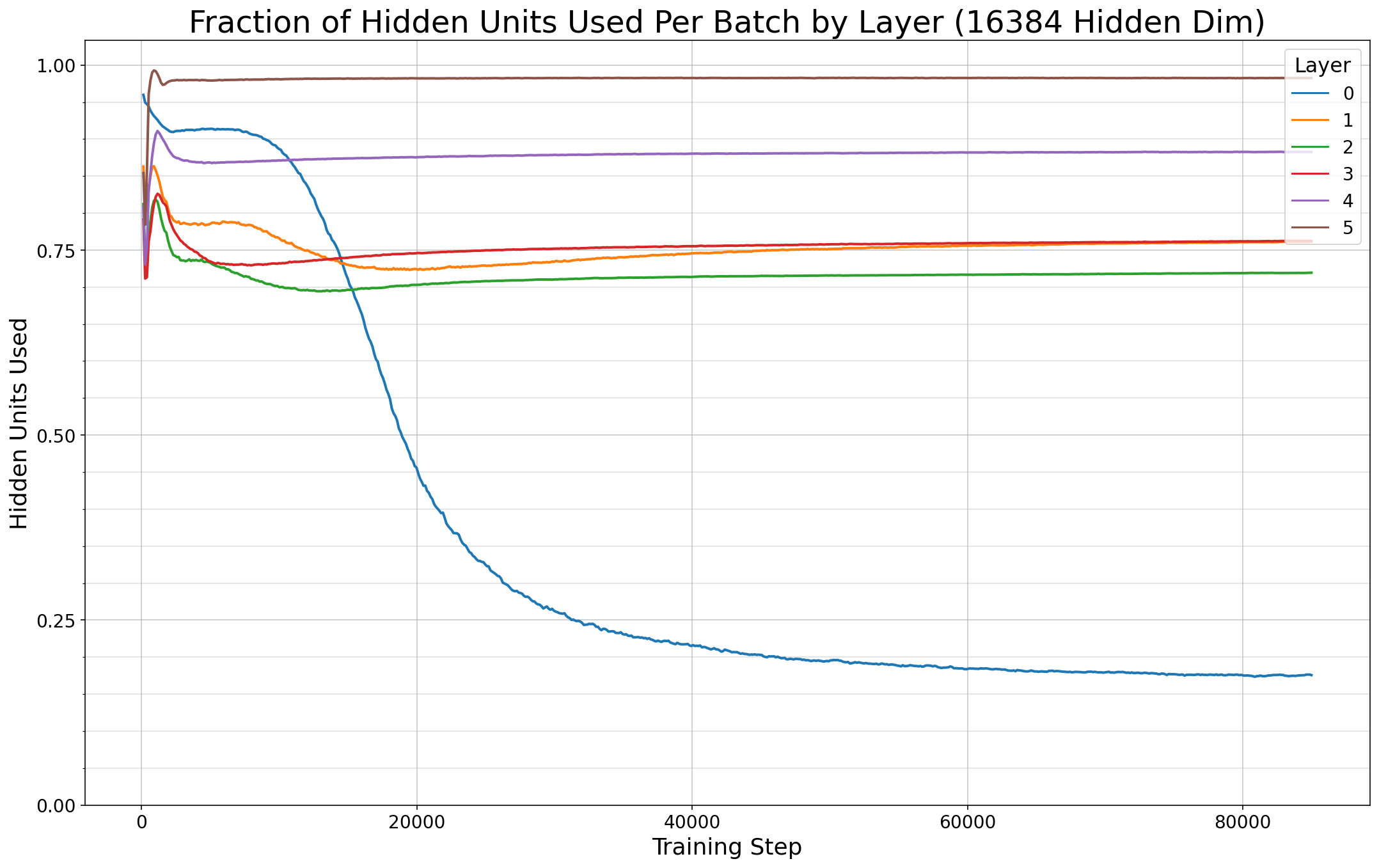}
    \caption{Fraction of available 16384 hidden units used per batch in a 6 layer model.}
    \label{per_batch_6l}
\end{figure}

\subsubsection{10 Layer}
\begin{figure}[H]
    \centering
    \includegraphics[width=0.8\linewidth]{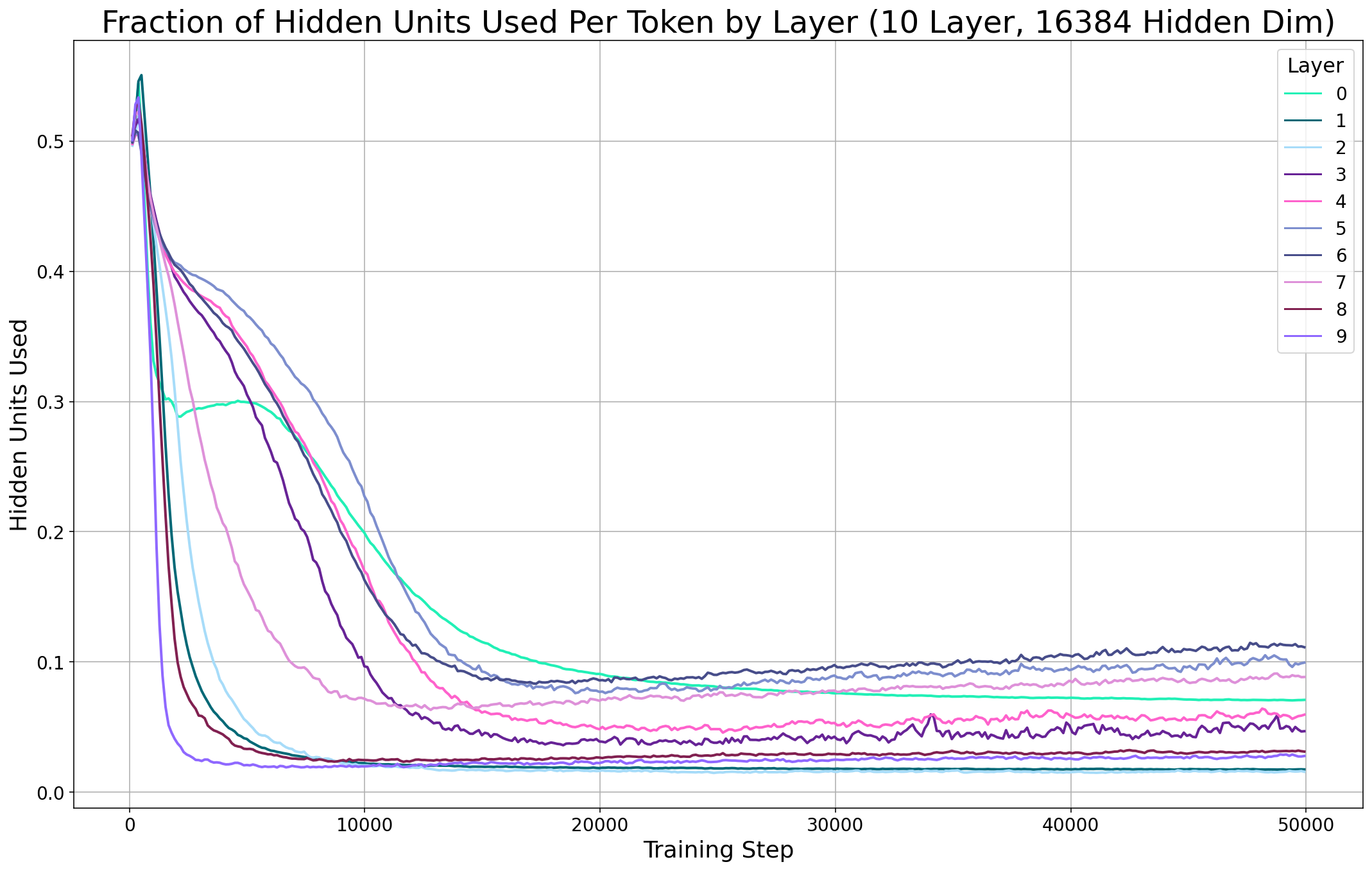}
    \caption{Fraction of available 16384 hidden units used per token in a 10 layer model.}
    \label{per_token_10l}
\end{figure}
\begin{figure}[H]
    \centering
    \includegraphics[width=0.8\linewidth]{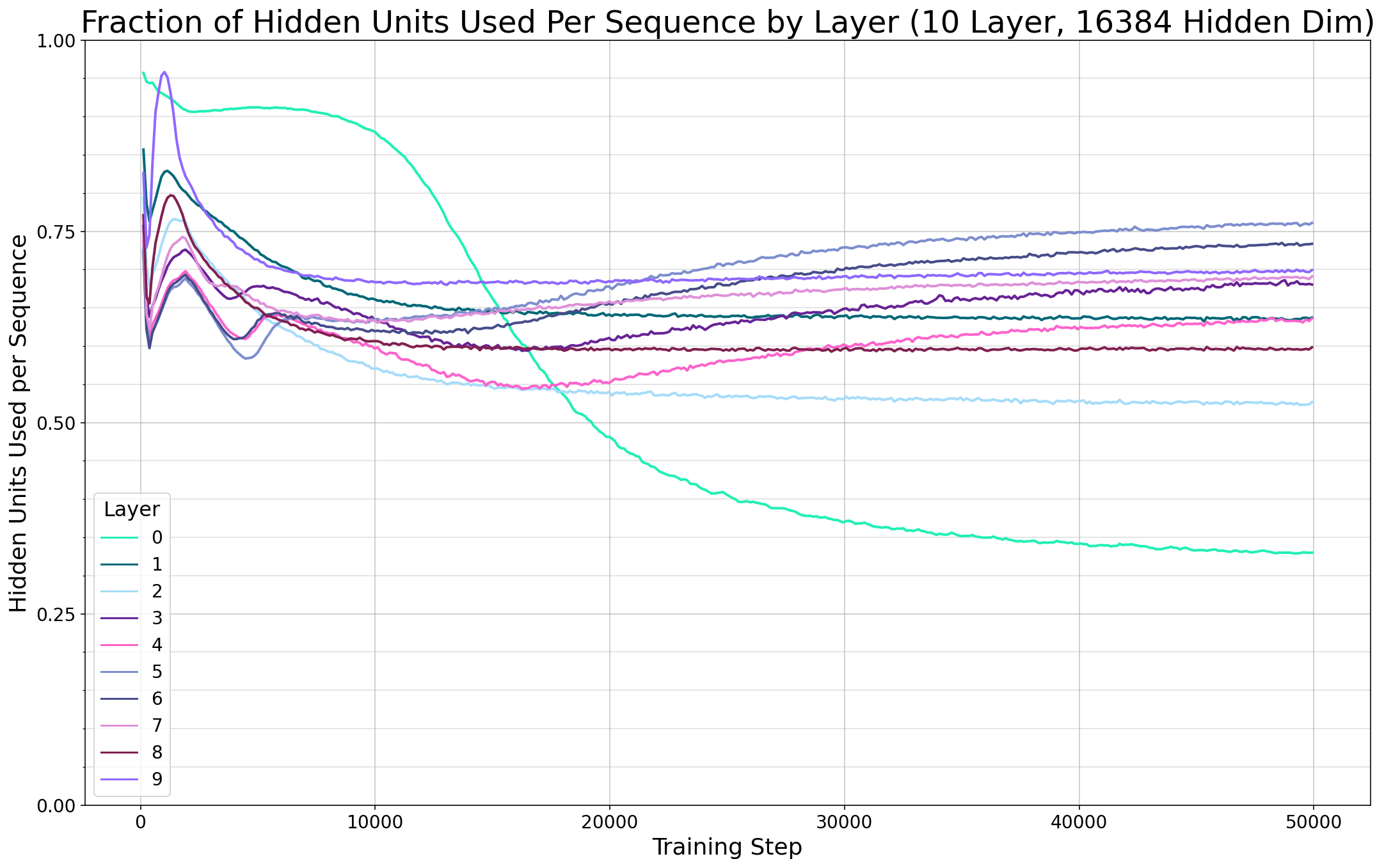}
    \caption{Fraction of available 16384 hidden units used per sequence in a 10 layer model.}
    \label{per_sequence_10l}
\end{figure}
\begin{figure}[H]
    \centering
    \includegraphics[width=0.8\linewidth]{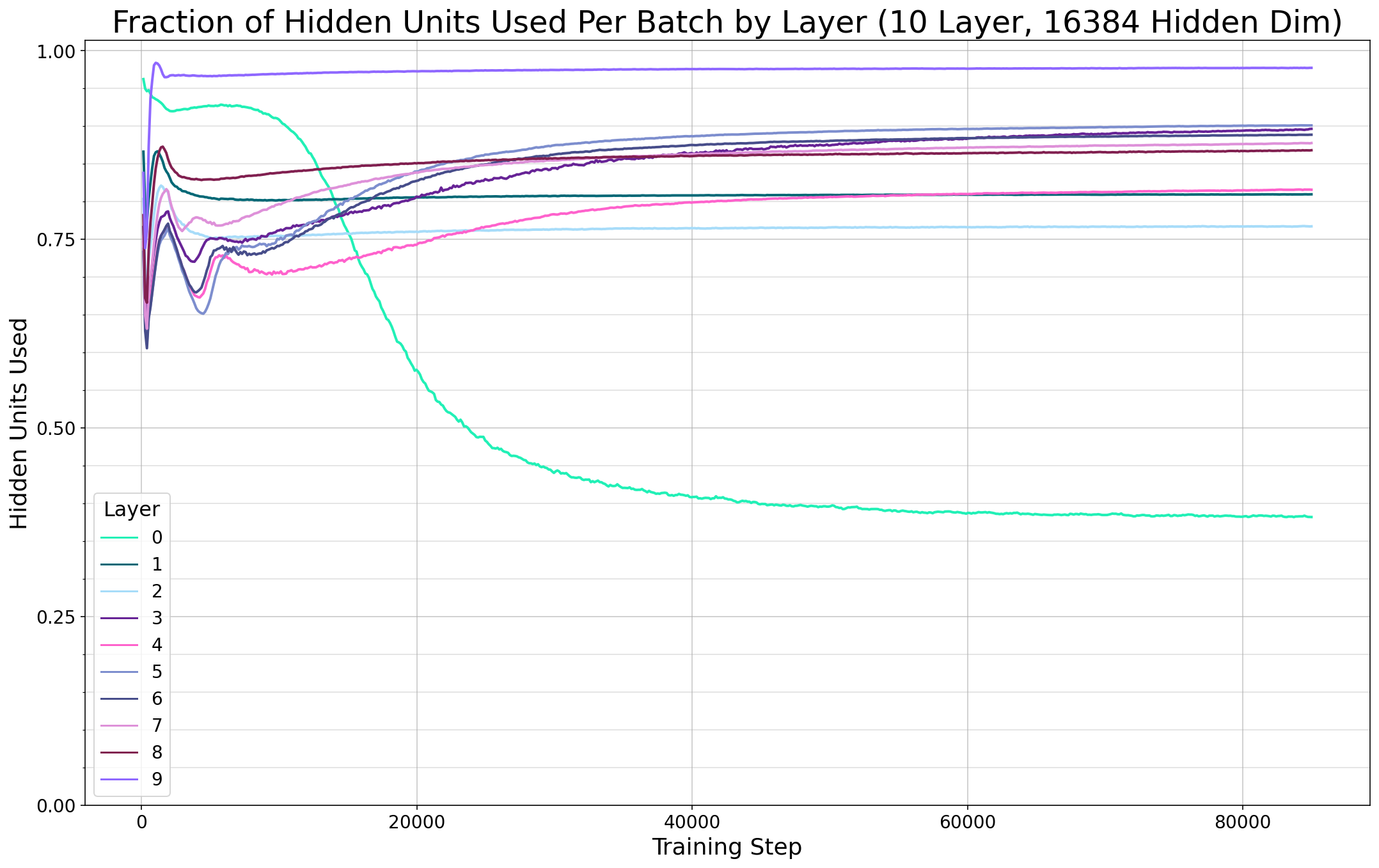}
    \caption{Fraction of available 16384 hidden units used per batch in a 10 layer model.}
    \label{per_batch_10l}
\end{figure}

\subsubsection{14 Layer}
\begin{figure}[H]
    \centering
    \includegraphics[width=0.8\linewidth]{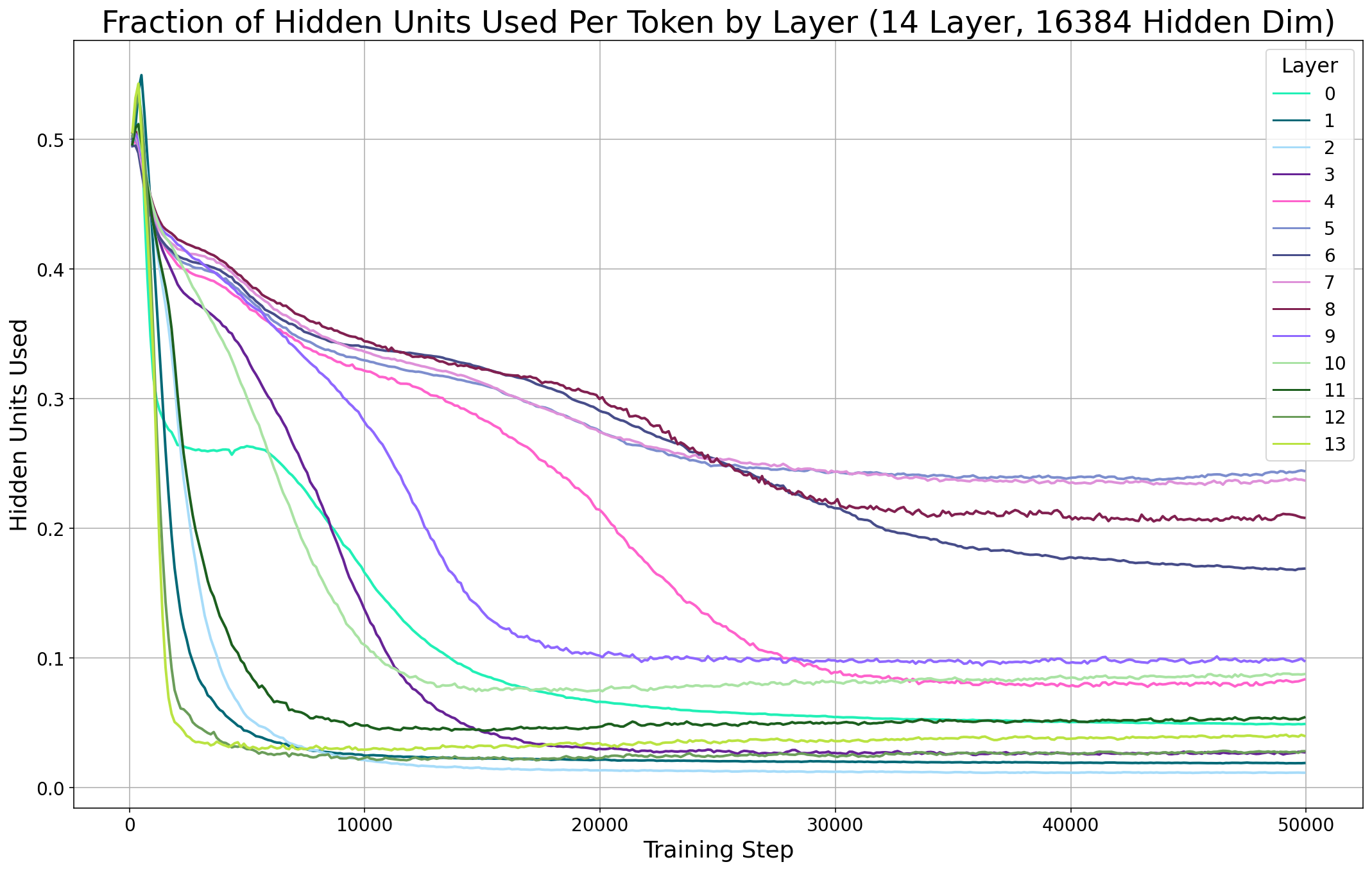}
    \caption{Fraction of available 16384 hidden units used per token in a 14 layer model.}
    \label{per_token_14l}
\end{figure}
\begin{figure}[H]
    \centering
    \includegraphics[width=0.8\linewidth]{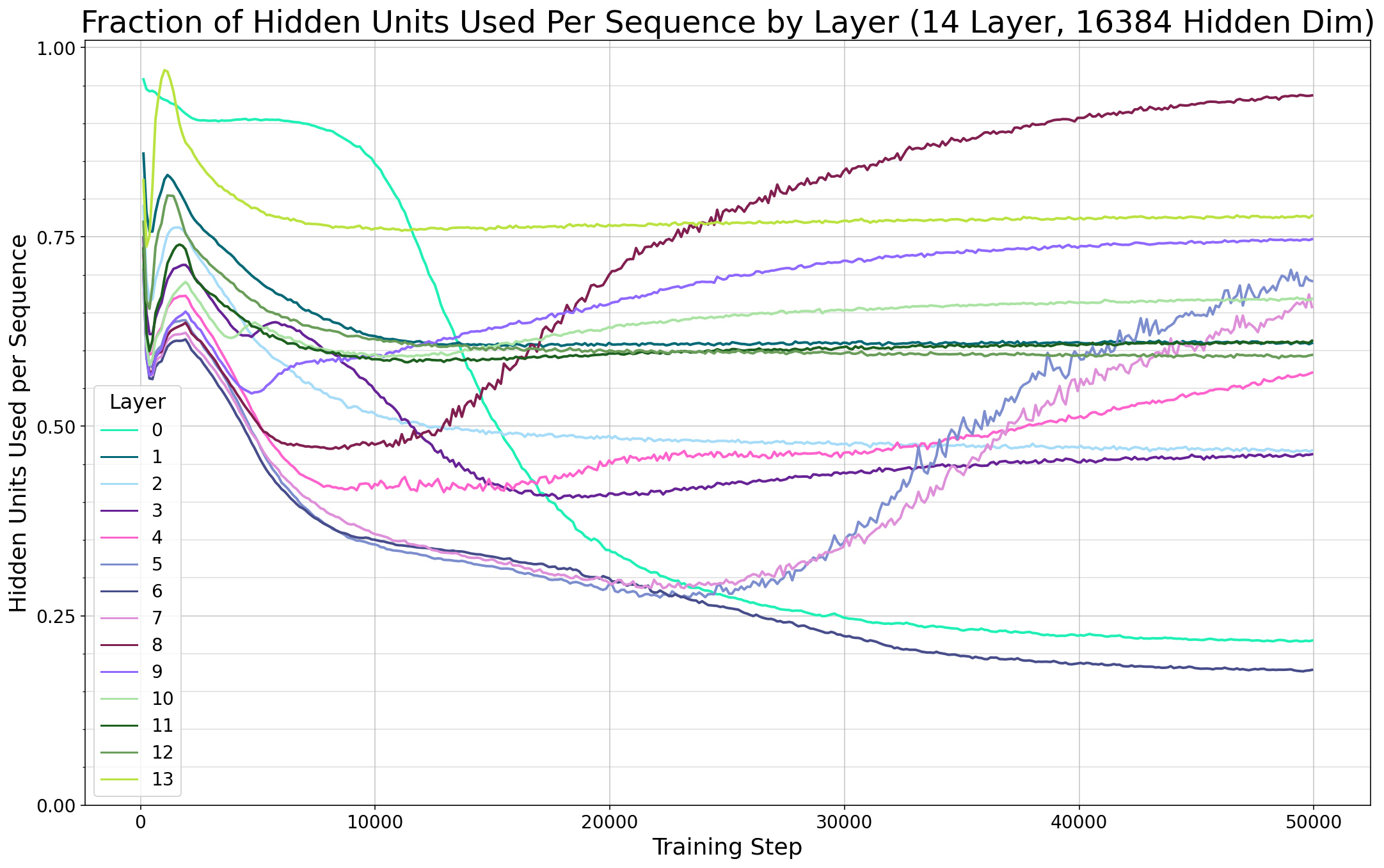}
    \caption{Fraction of available 16384 hidden units used per sequence in a 14 layer model.}
    \label{per_sequence_14l}
\end{figure}
\begin{figure}[H]
    \centering
    \includegraphics[width=0.8\linewidth]{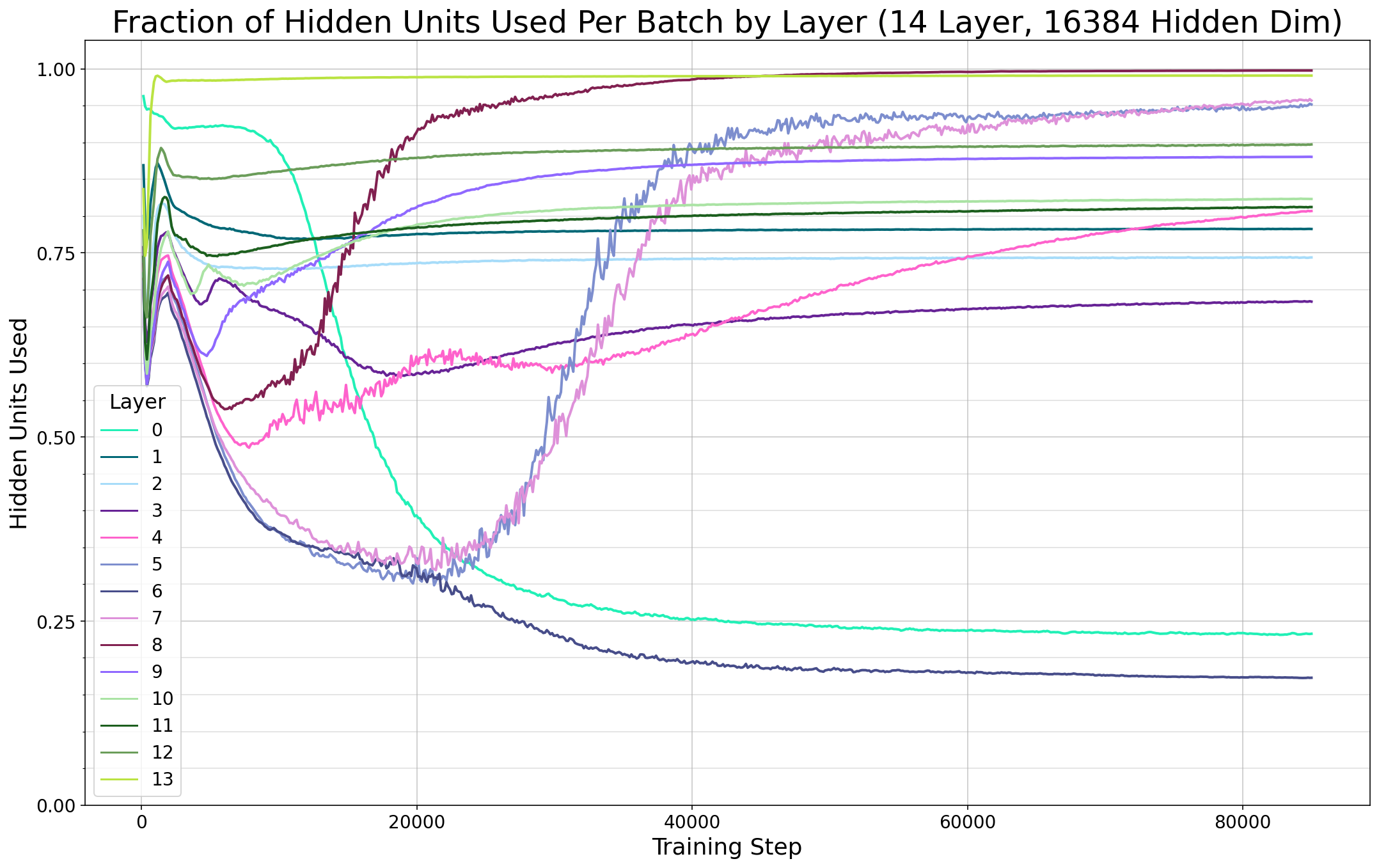}
    \caption{Fraction of available 16384 hidden units used per batch in a 14 layer model.}
    \label{per_batch_14l}
\end{figure}

\subsection{Sensitivity to Hidden Dim Size Plots}
\label{hidden_dim_sensitivity}
This section shows token, sequence, and batch hidden unit use for 6 layer models of hidden dimensions 16384 and 8192 (32768 is the default for the main paper). 

\subsubsection{Hidden Dimension: 16384}
\begin{figure}[H]
    \centering
    \includegraphics[width=0.8\linewidth]{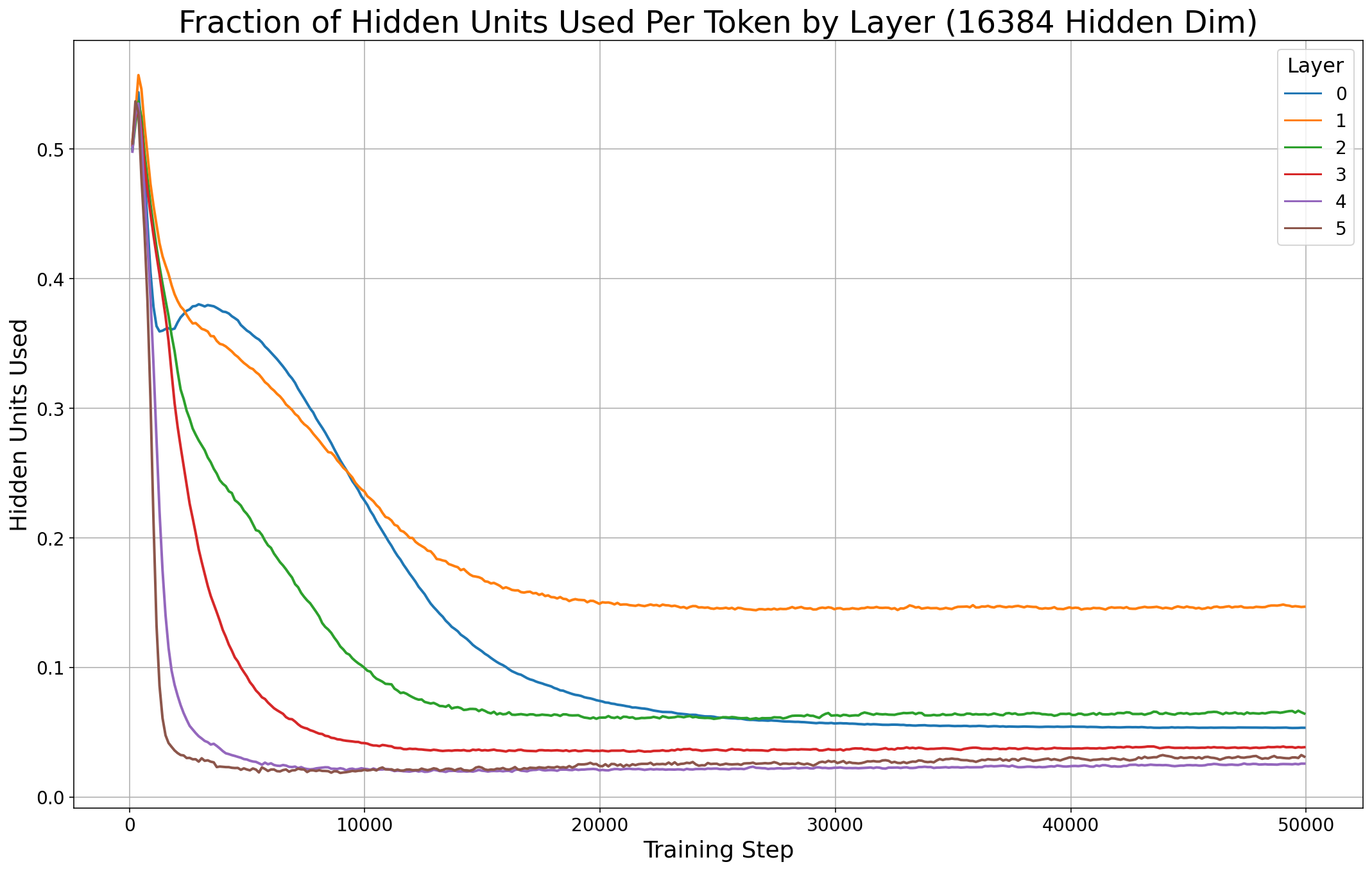}
    \caption{Fraction of available 16384 hidden units used per batch in a 6 layer model.}
    \label{per_token_6l_16384}
\end{figure}
\begin{figure}[H]
    \centering
    \includegraphics[width=0.8\linewidth]{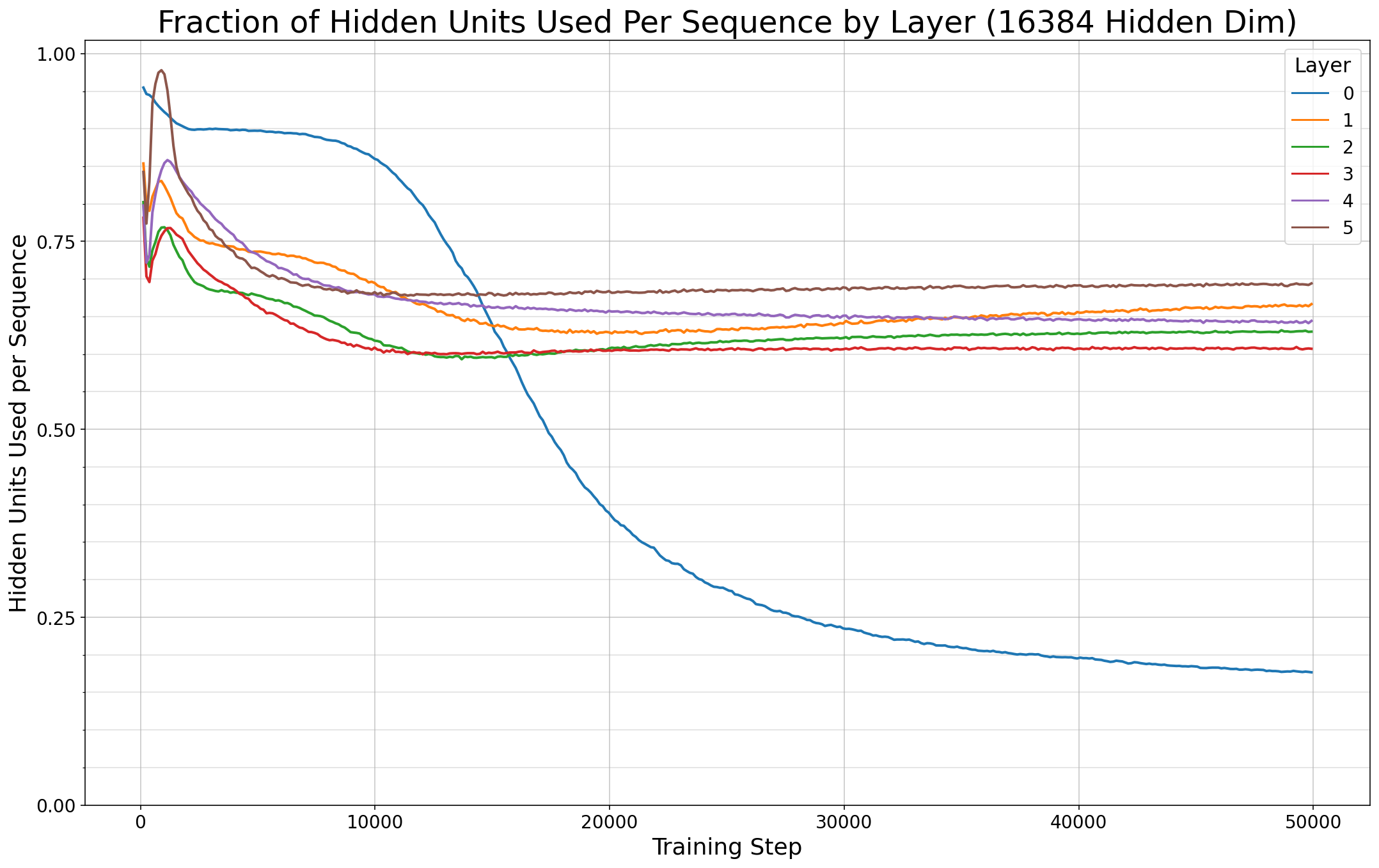}
    \caption{Fraction of available 16384 hidden units used per sequence in a 6 layer model.}
    \label{per_sequence_6l_16384}
\end{figure}

\begin{figure}[H]
\centering
    \includegraphics[width=0.8\linewidth]{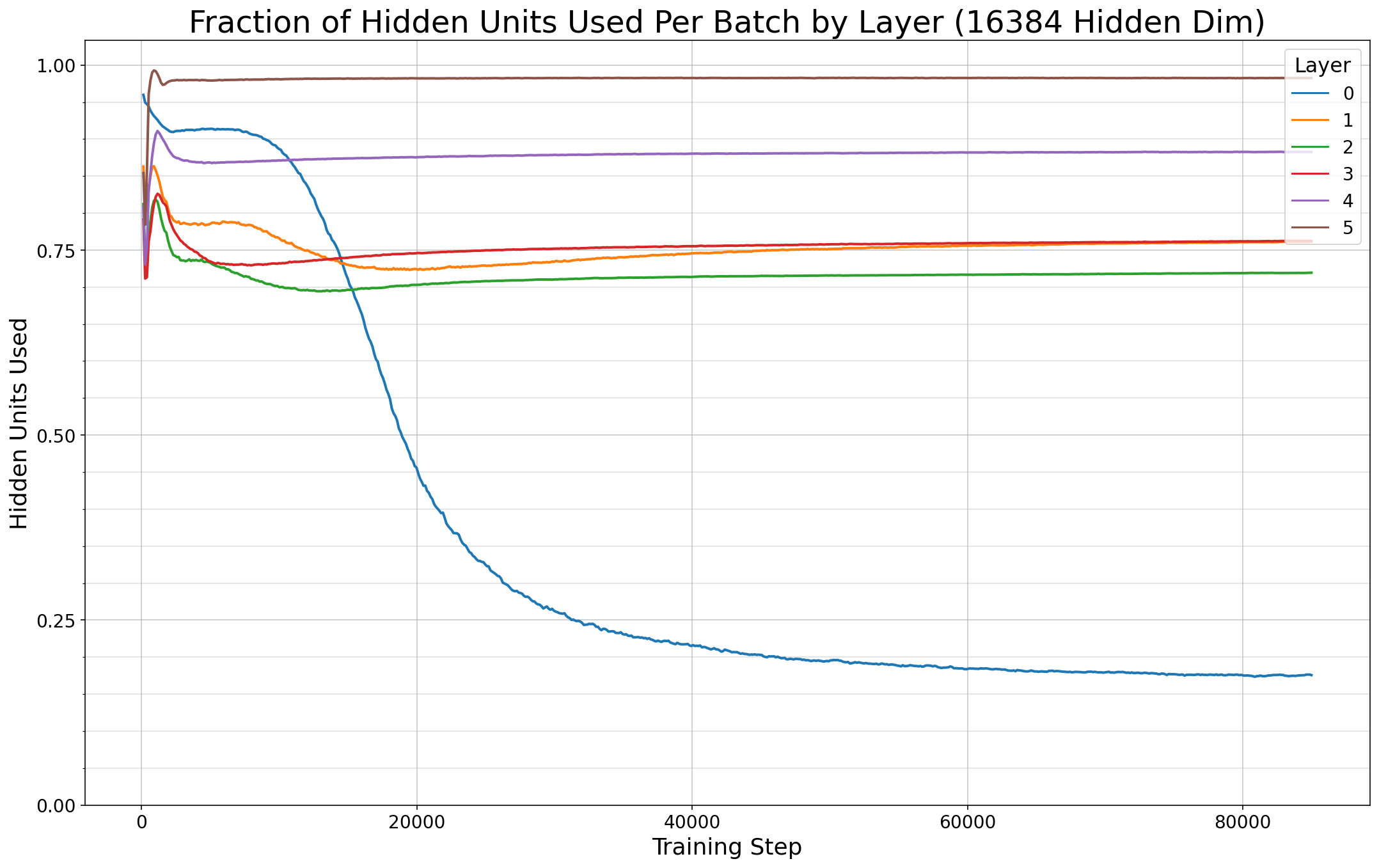}
    \caption{Fraction of available 16384 hidden units used per batch in a 6 layer model.}
    \label{per_batch_6l_16384}
\end{figure}

\subsubsection{Hidden Dimension: 8192}
\begin{figure}[H]
\centering
    \includegraphics[width=0.8\linewidth]{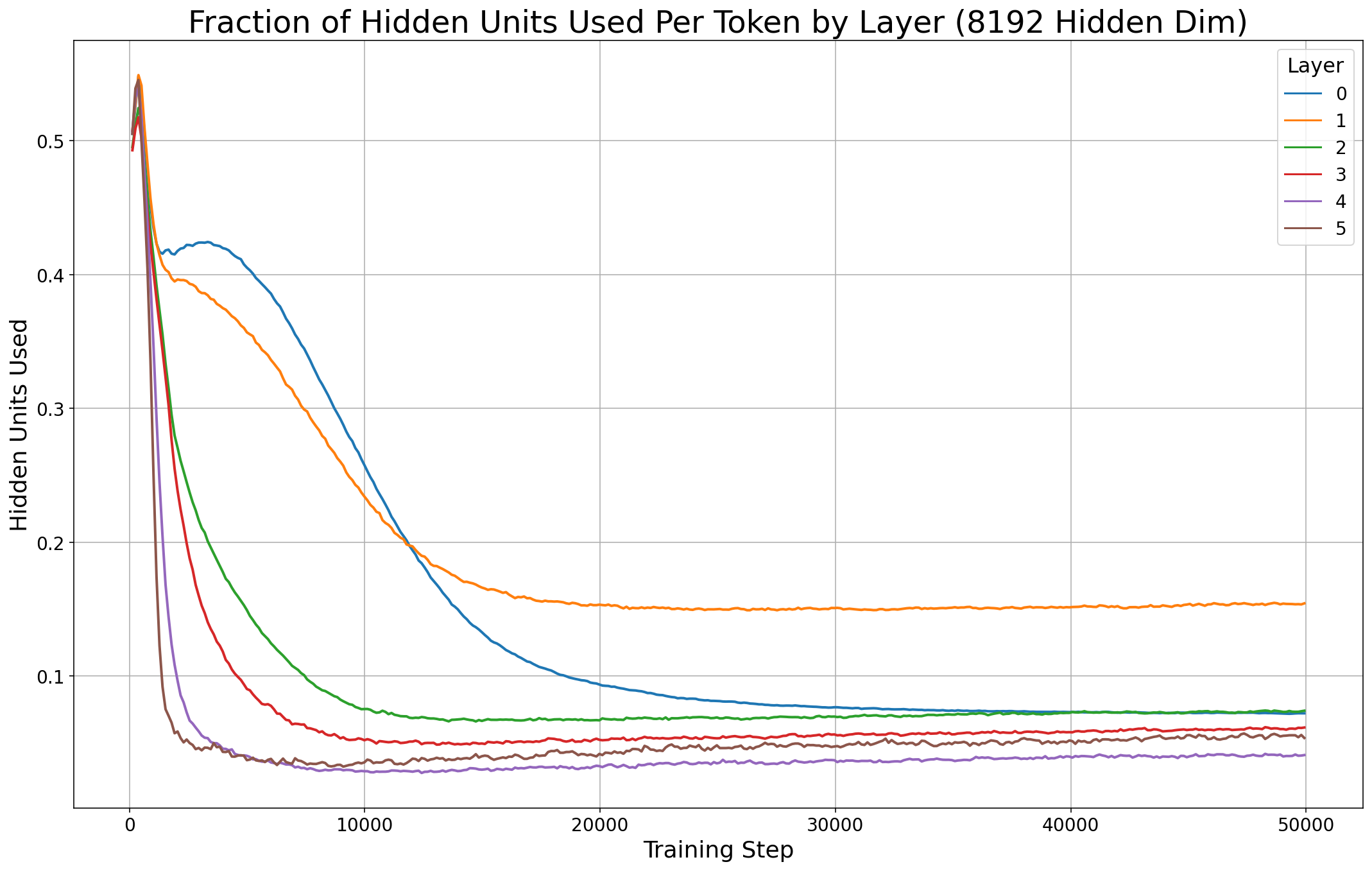}
    \caption{Fraction of available 8192 hidden units used per token in a 6 layer model.}
    \label{per_token_6l_8192}
\end{figure}
\begin{figure}[H]
    \centering
    \includegraphics[width=0.8\linewidth]{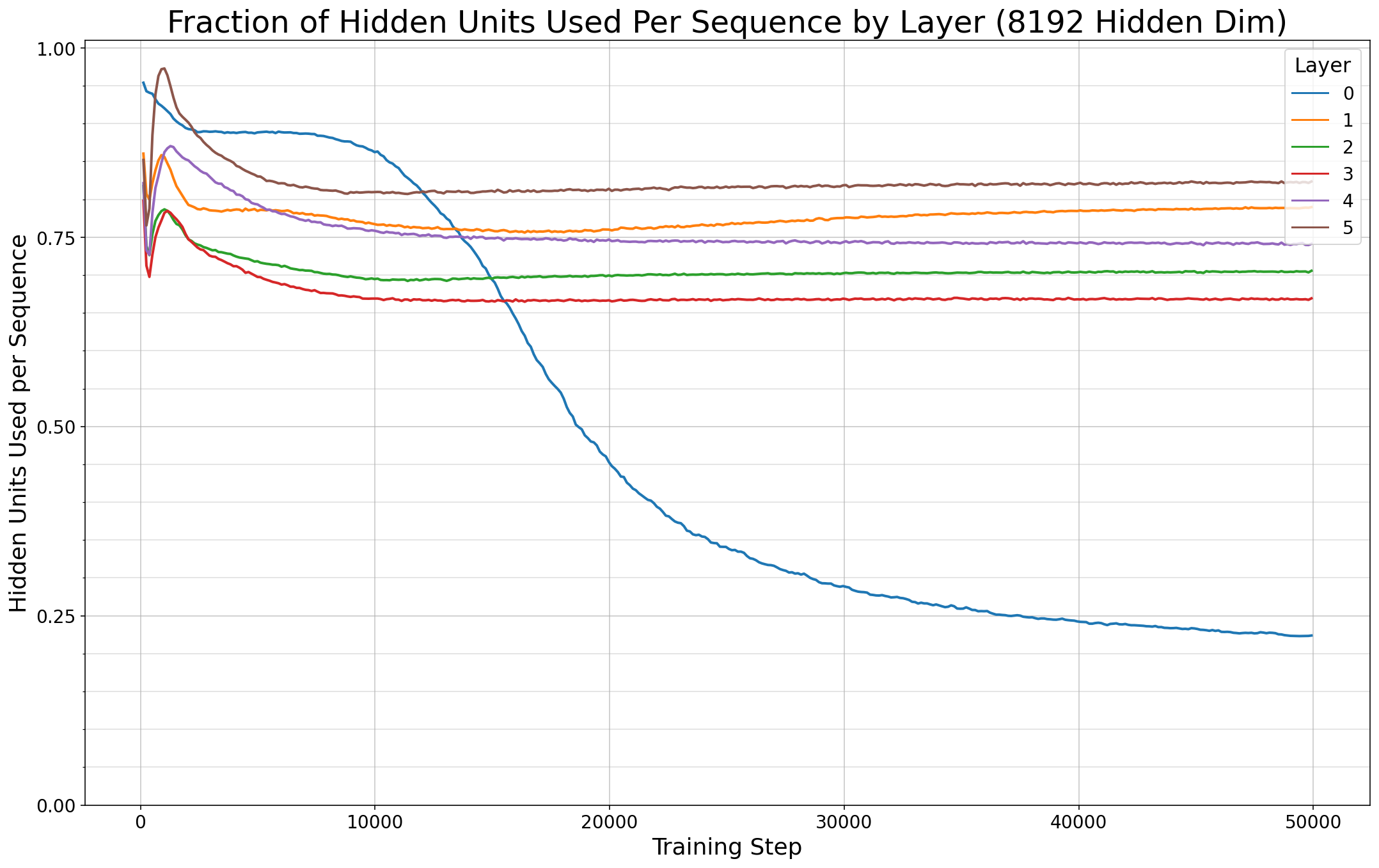}
    \caption{Fraction of available 8192 hidden units used per sequence in a 6 layer model.}
    \label{per_sequence_6l_8192}
\end{figure}
\begin{figure}[H]
    \centering
    \includegraphics[width=0.8\linewidth]{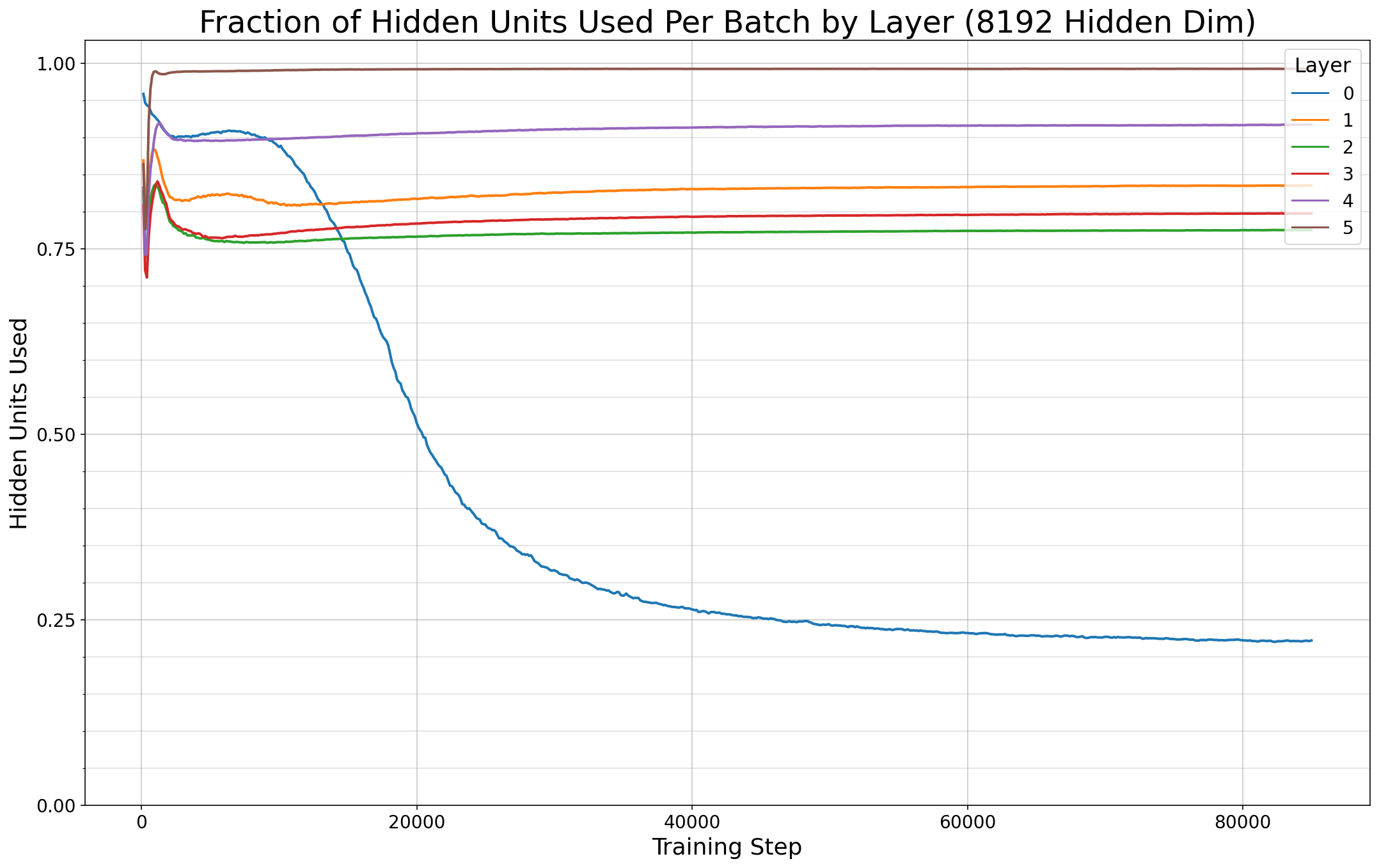}
    \caption{Fraction of available 8192 hidden units used per batch in a 6 layer model.}
    \label{per_batch_6l_8192}
\end{figure}

\subsection{Sensitivity to Learning Rate Plots}
This section shows token, sequence, and batch hidden unit use for models using a max learning rate of 1e-3 and 2e-3 (3e-3 is the standard used in the main paper)
\label{learning_rate_sensitivity}
\subsubsection{Max LR: 1e-3}
\begin{figure}[H]
    \centering
    \includegraphics[width=0.8\linewidth]{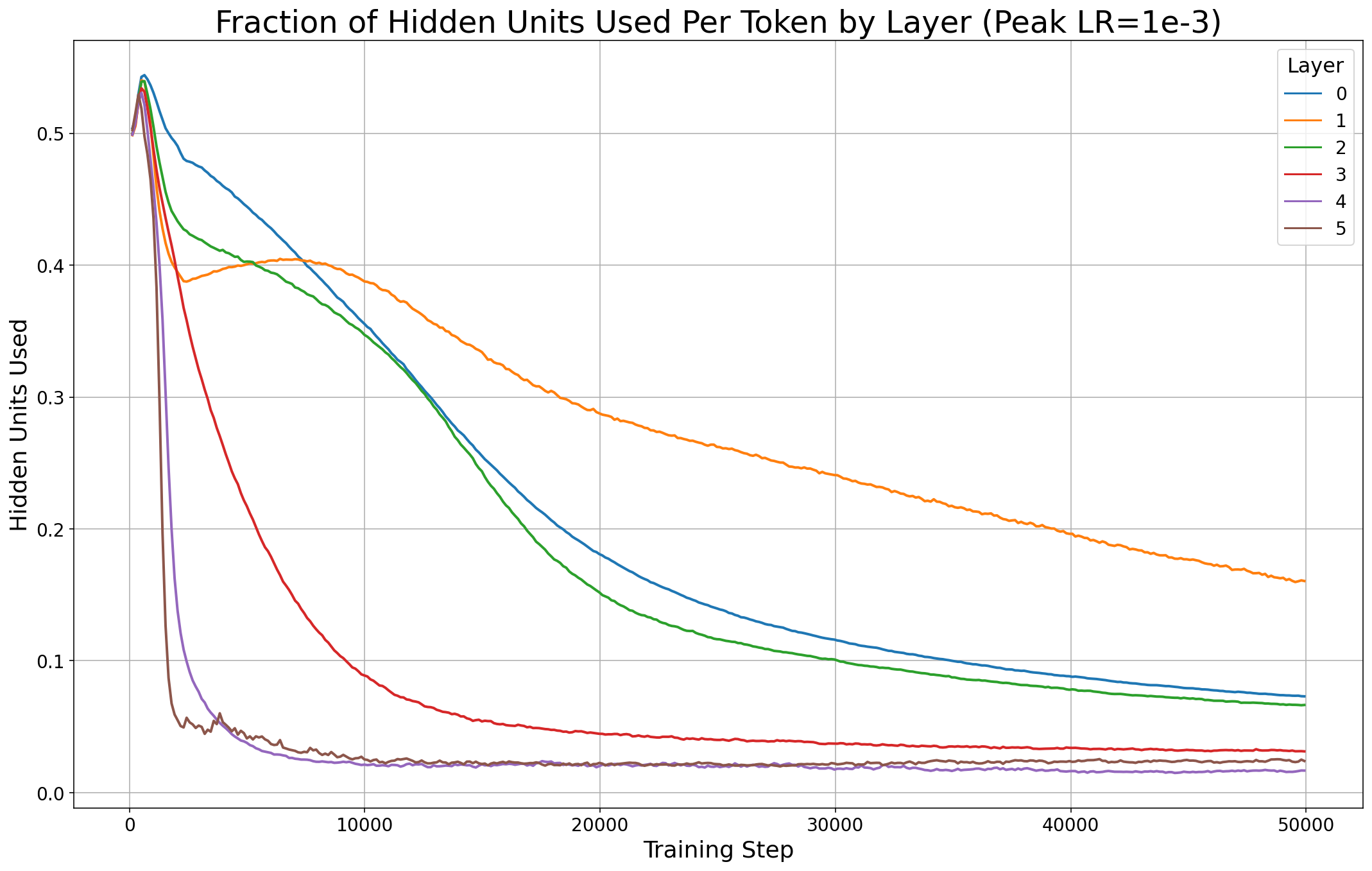}
    \caption{Fraction of available 32768 hidden units used per token in a model with a peak LR of 1e-3.}
    \label{per_token_6l_1e3}
\end{figure}
\begin{figure}[H]
    \centering
    \includegraphics[width=0.8\linewidth]{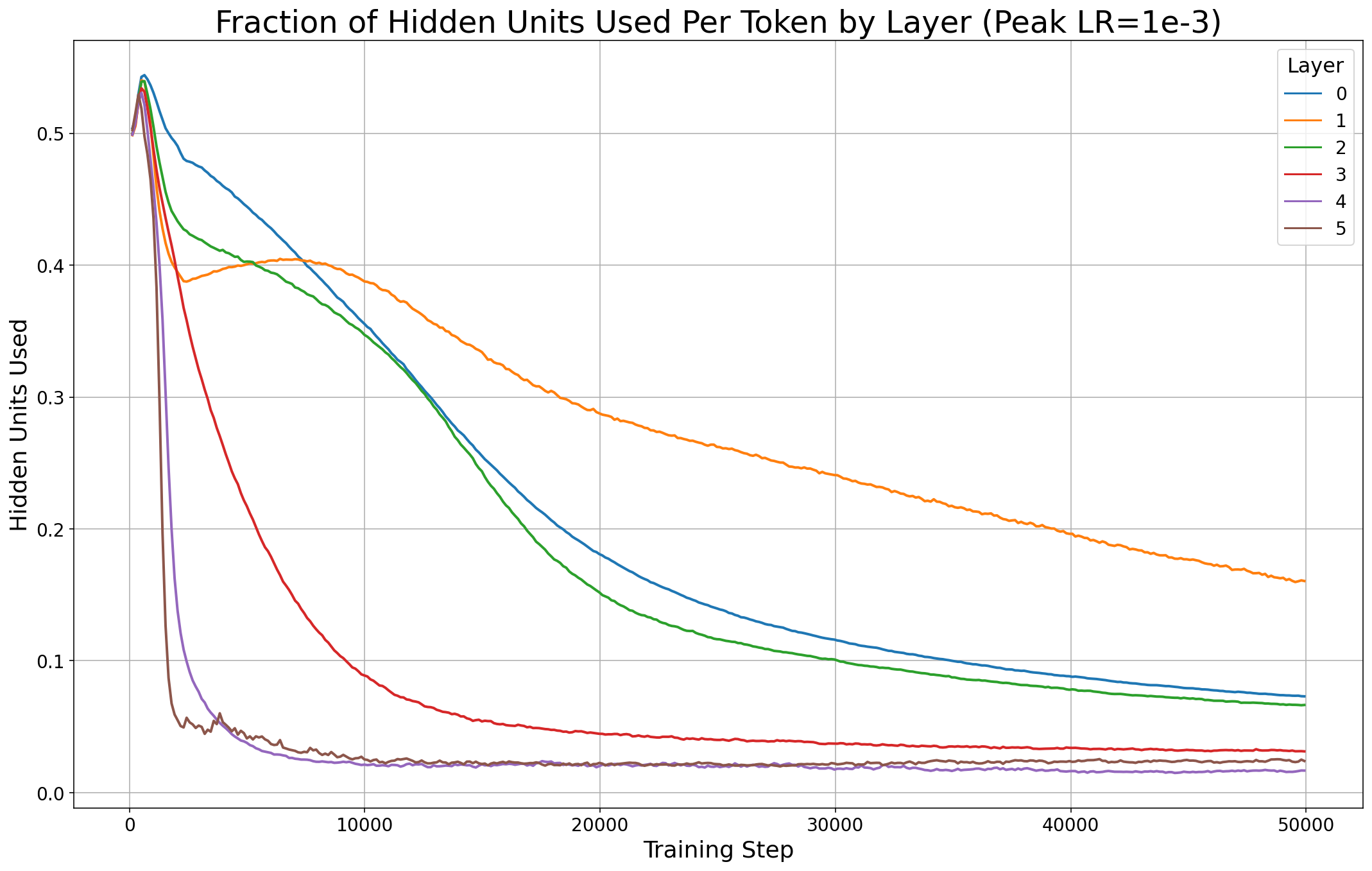}
    \caption{Fraction of available 32768 hidden units used per sequence in a model with a peak LR of 1e-3.}
    \label{per_sequence_6l_1e3}
\end{figure}
\begin{figure}[H]
    \centering
    \includegraphics[width=0.8\linewidth]{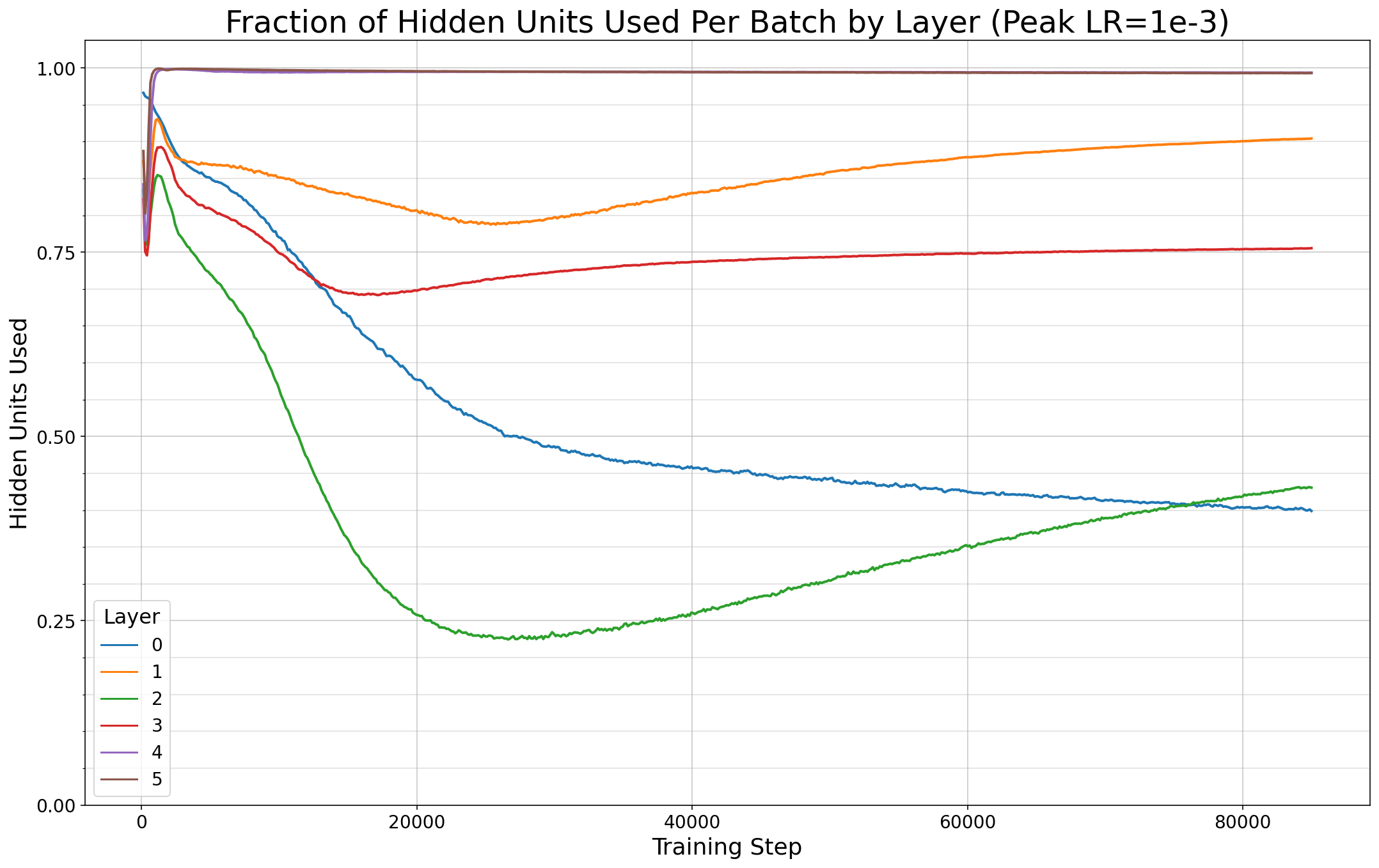}
    \caption{Fraction of available 32768 hidden units used per batch in a model with a peak LR of 1e-3.}
    \label{per_batch_6l_1e3}
\end{figure}

\subsubsection{Max LR: 2e-3}
\begin{figure}[H]
    \centering
    \includegraphics[width=0.8\linewidth]{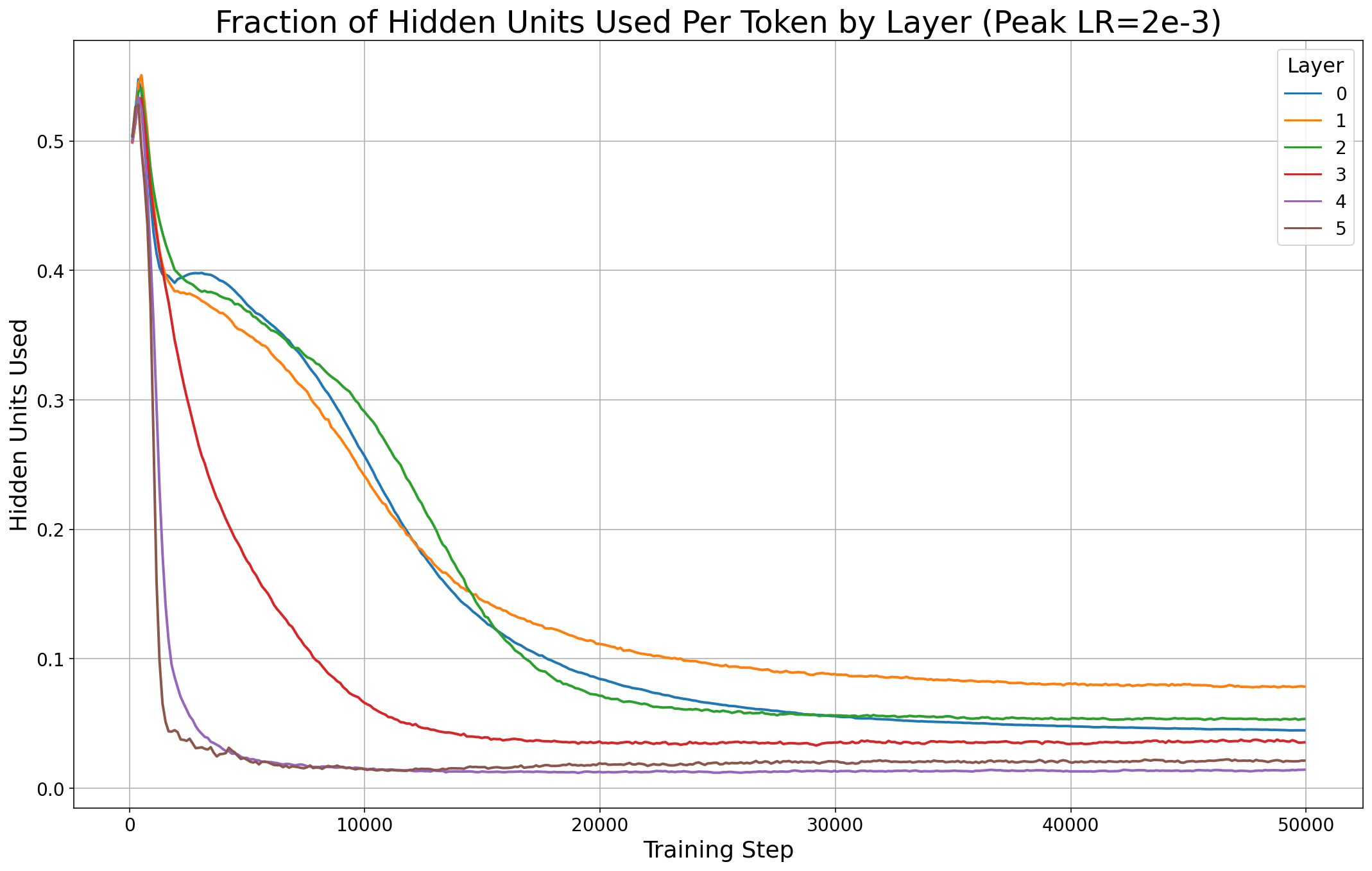}
    \caption{Fraction of available 32768 hidden units used per token in a model with a peak LR of 2e-3.}
    \label{per_token_6l_2e3}
    \end{figure}
\begin{figure}[H]
    \centering
    \includegraphics[width=0.8\linewidth]{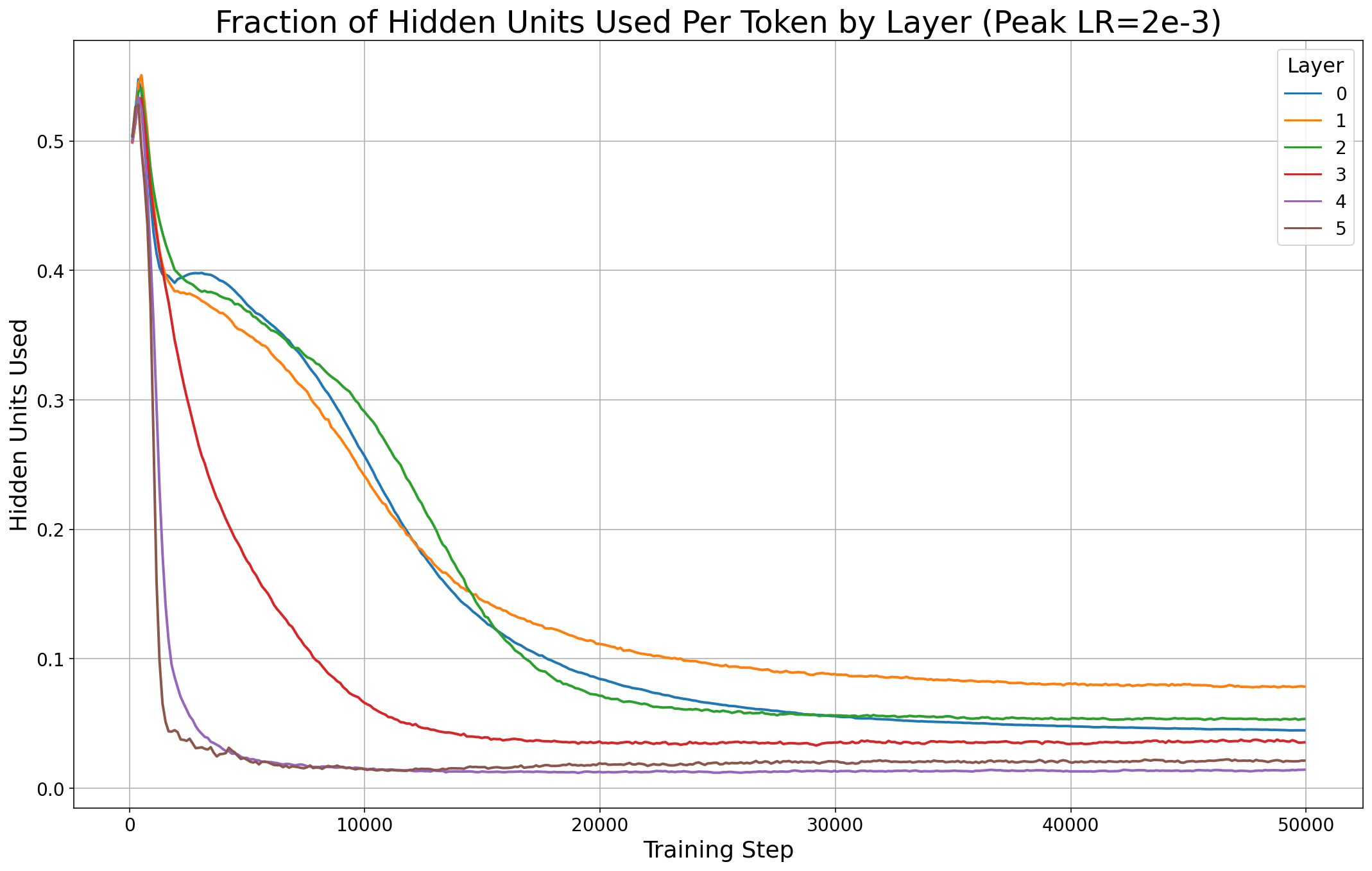}
    \caption{Fraction of available 32768 hidden units used per sequence in a model with a peak LR of 2e-3.}
    \label{per_sequence_6l_2e3}
\end{figure}

\begin{figure}[H]
    \centering
    \includegraphics[width=0.8\linewidth]{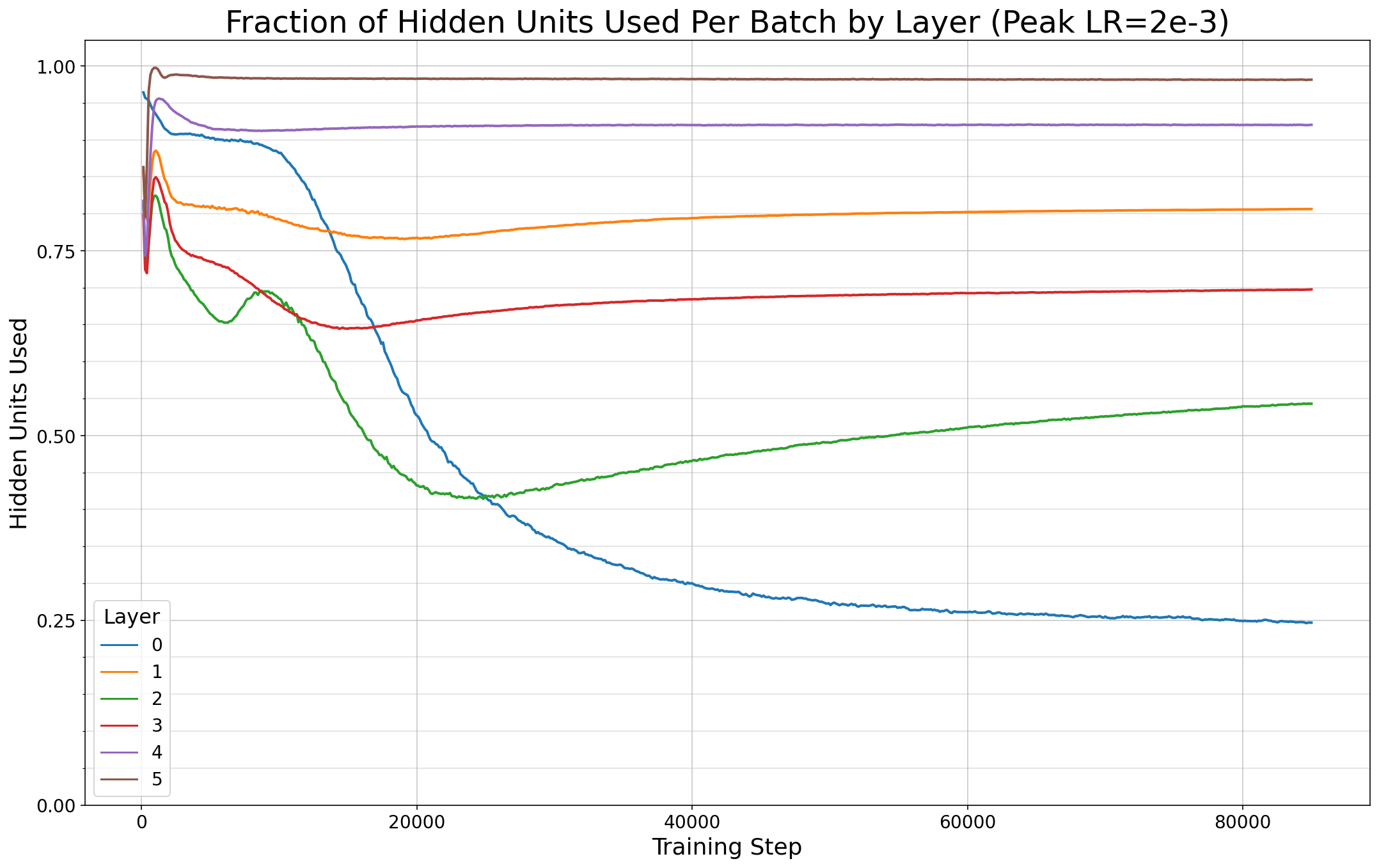}
    \caption{Fraction of available 32768 hidden units used per batch in a model with a peak LR of 2e-3.}
    \label{per_batch_6l_2e3}
\end{figure}

\subsection{Metrics Definition Pseudocode}
\label{code_definitions}
In the code samples that follow,`batch` is an array of hidden state activations of size (batch, sequence, hidden), and `seq` is an array of hidden state activations of size (sequence, hidden). When a metric over sequences is shown, it is calculated by mapping a sequence-level function over the sequences in a batch, and then averaging. 
\subsubsection{Percentile Metrics}
\lstset{language=Python}
\begin{lstlisting}
def get_rescaled_percentile(fract_nonzero: array, desired_percentile: int
):
    fract_zero = 1 - fract_nonzero
    rescaled_percentile = desired_percentile * fract_nonzero + 100 * fract_zero
    return rescaled_percentile
\end{lstlisting}

\begin{lstlisting}
def percentile_used_dimension_count(seq: array, percentile: int
) -> array:

    seq_dim = seq.shape[0]
    use_count_by_dimension = sum(seq, axis=0)
    frac_dimensions_gt_0 = (
      sum(use_count_by_dimension > 0) / use_count_by_dimension.shape[0]
    )
    new_percentile = get_rescaled_percentile(frac_dimensions_gt_0, percentile)
    return percentile(use_count_by_dimension, new_percentile) / seq_dim
\end{lstlisting}

\subsubsection{Hidden Unit Use Metrics} 
\begin{lstlisting}
def sequence_total_dimensions_used(seq: array) -> array:
    summed_activations_over_sequence = sum(seq, axis=0)
    return sum((summed_activations_over_sequence > 0).astype(int32))
\end{lstlisting}

\begin{lstlisting}
def flatten(batch: array) -> array:
    return batch.reshape(batch.shape[0] * batch.shape[1], batch.shape[2])
\end{lstlisting}
\begin{lstlisting}
def total_batch_dimensions_used(batch: array) -> array:
    return sequence_total_dimensions_used(flatten(batch))
\end{lstlisting}

\end{document}